\documentclass[10pt,twocolumn,letterpaper]{article}            
\usepackage[pagenumbers]{cvpr} 
\usepackage[dvipsnames]{xcolor}
\usepackage{multirow}
\usepackage{soul}
\usepackage{wrapfig}
\usepackage{epsfig}
\usepackage{caption}
\usepackage{adjustbox}
\usepackage[symbol]{footmisc}
\usepackage[accsupp]{axessibility}

\definecolor{cvprblue}{rgb}{0.21,0.49,0.74}
\usepackage[pagebackref,breaklinks,colorlinks,citecolor=cvprblue]{hyperref}

\def\paperID{22} 
\def\confName{CVPR}
\def\confYear{2024}

\title{Probing Conceptual Understanding of Large Visual-Language Models}

\author{Madeline Schiappa\textsuperscript{1} \qquad \qquad Raiyaan Abdullah\textsuperscript{1*} \qquad \qquad Shehreen Azad\textsuperscript{1} \\
Jared Claypoole\textsuperscript{2} \qquad \qquad Michael Cogswell\textsuperscript{2} \qquad \qquad Ajay Divakaran\textsuperscript{2}\\
Yogesh Rawat\textsuperscript{1}\\
\textsuperscript{1}Center for Research in Computer Vision, University of Central Florida\\
\textsuperscript{2}SRI International
\\ 
}
\begin{document}
\twocolumn[{
\maketitle
\begin{center}
    \centering 
    \captionsetup{type=figure}
    \includegraphics[width=\textwidth]{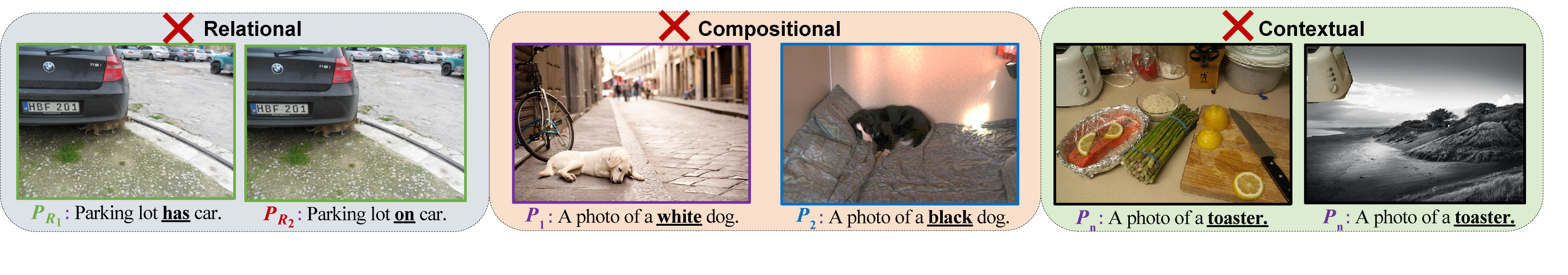}
    \caption{\textbf{Conceptual understanding of an existing V+L model.} Here, CLIP failure to understand relational, compositional and contextual reasoning is shown. This benchmark presents three datasets to evaluate V+L models on relational, compositional, and contextual understanding. They utilize image-text matching tasks with predicate, object/subject, compositions, or background swaps.}
\label{teaser}
\end{center}
}]

\begin{abstract}
In recent years large visual-language (V+L) models have achieved great success in various downstream tasks. However, it is not well studied whether these models have a conceptual grasp of the visual content. In this work we focus on conceptual understanding of these large V+L models. To facilitate this study, we propose novel benchmarking datasets for probing three different aspects of content understanding, 1) \textit{relations}, 2) \textit{composition}, and 3) \textit{context}. Our probes are grounded in cognitive science and help determine if a V+L model can, for example, determine if snow garnished with a man is implausible, or if it can identify beach furniture by knowing it is located on a beach. We experimented with many recent state-of-the-art V+L models and observe that these models mostly \textit{fail to demonstrate} a conceptual understanding. This study reveals several interesting insights such as that \textit{cross-attention} helps learning conceptual understanding, and that CNNs are better with \textit{texture and patterns}, while Transformers are better at \textit{color and shape}. We further utilize some of these insights and investigate a \textit{simple finetuning technique} that rewards the three conceptual understanding measures with promising initial results. The proposed benchmarks will drive the community to delve deeper into conceptual understanding and foster advancements in the capabilities of large V+L models. The code and dataset is available at: \url{https://tinyurl.com/vlm-robustness}
\end{abstract}
\footnotetext{*Corresponding Author: raiyaanabdullah@gmail.com}

\section{Introduction}

Humans navigate the world by learning an ``understanding'' of how it works.  Understanding may be defined as the underlying organization of all concepts, including objects, situations, events, and more \cite{committee2015psychological,mayer1989models}. They are organized in our brains as \textit{conceptual maps}, which encode structured, relational information \cite{annurev_frankland_greene}. \textit{Conceptual maps} highlight major objects and actions in a system and the causal relations between them. While deep learning models have impressive performance in a variety of tasks, it is still unclear if their impressive performance is due to learnt \textit{conceptual maps}.

\noindent
Large visual-language (V+L) models are recently and greatly successful deep learning models that learn representations of image and text in a shared space. These representations are useful for downstream tasks like image classification, visual-question answering, image retrieval and more \cite{zhou2022conditional,Wang_2022_CVPR,Rao_2022_CVPR,clip,xu2022bridge,alayrac2022flamingo}. However, for use in real-world applications, it is also vital that models ``understand'' rather than memorize to perform on more general tasks 
\cite{piloto2022intuitive}. While large-language models have been shown to have a moderate amount of ``theory of mind,'' as measured by conceptual consistency \cite{sahu2022unpacking}, V+L models have not been investigated in a similar way using real-world examples. This is partly because images are more challenging, as shown by preliminary studies \cite{thrush2022winoground,diwan2022winoground,carlini2023extracting}. With this in mind, we focus on probing models on their conceptual maps.

\noindent
We develop a benchmark by combining insights from well-known tests such as the Peabody Picture test, semantic analysis underpinning knowledge bases such as ConceptNet \cite{speer2017conceptnet}, and comprehension in elementary school education \cite{sahu2022unpacking} to identify three key areas for probing: relations, composition, and context (Figure \ref{teaser}). Our benchmark could be seen as a computational instantiation of visual comprehension testing along three important fundamental skills. These skills form a compact set of necessary, but not sufficient, prerequisites for key tasks such as concept transfer, analysis, evaluation, and generation. They thus provide us a basis for probing comprehension of large V+L models.

\noindent
We propose three benchmark datasets, \textit{Probe-R}, \textit{Probe-C}, and \textit{Probe-B}.
Probe-R looks at model understanding of possible object relations by comparing an image to a correct prompt and an incorrect prompt where the predicate is swapped with an unlikely relation. Probe-C looks at model understanding of possible compositional relations by comparing two images and two prompts where either the composition is swapped with an antonym or the object is swapped. Finally, Probe-B looks at model understanding of objects and their relationships to their surroundings by removing background and observing the change in performance.

\noindent
We experimented with several state-of-the-art V+L models and provide several interesting insights regarding these models. 
For compositional understanding, we observe that (1) models struggle with compositionality, and (2) CNN based backbones may be better at recognizing texture and patterns while ViT backbones are better with color and shape. For relational understanding, we observe that (1) both modality specific attention and co-attention in parallel improve relational understanding, and (2) Predicate swapping that violates expectations surfaces the lack of an underlying conceptual model. For contextual understanding we observe that (1) models tend to not use context in order to recognize most objects, again indicating a lack of an underlying conceptual model. We further utilize these findings and develop a simple finetuning approach based on selective negatives paradigm and observe improvement on our understanding-related probes.

\noindent
In summary, we make the following contributions:

\begin{itemize}
\setlength
    \item{We study the capability of existing large V+L models for complex visual perception focusing on relational, compositional, and contextual understanding. }
    \item{We propose three benchmark datasets: Probe-R, Probe-C, and Probe-B focusing on subject-object relations, composition-object relations, and background-object relations.}
    \item{We perform extensive evaluation of existing models and provide new insights about their  capabilities.}
    \item{We present a simple approach, based on prompting that rewards compositionality and preservation of relations between objects, which yields a more robust performance on complex visual perception tasks.}
\end{itemize}

\section{Related Works}
\label{sec:relwork}
Several works have probed models to understand what models are learning \cite{thrush2022winoground,diwan2022winoground,yuksekgonul2022and,patterson2016coco,Johnson_2017_CVPR,vlchecklist,aro,svlc,controlledimcaps,crepe,sugarcrepe}. Table \ref{tab:comparison} shows a comparison of our proposed benchmark against several other existing works that probe the different understanding property of V+L models. From the table, it is evident that none of the existing works probe the contextual understanding of V+L models, which our proposed dataset does. Moreover, our proposed benchmark has more images and is evaluated on more models than most of the existing methods, which will be discussed later. Even though SVLC \cite{svlc} is a large-scale benchmark in terms of both size of dataset and number of evaluation models than our proposed benchmark, this is not sufficient for contextual understanding of the V+L models. 
An extension of Winoground  \cite{diwan2022winoground} showed that models perform worse than humans because it requires both compositional understanding and commonsense reasoning. Without disentangling the individual skills required to perform well, it limits insights to why/how they are failing and where to improve. In this work, we generate a benchmark that isolates components of understanding into compositional, relational, and context, allowing for more detailed insights.  

\begin{figure*}[t!]
    \centering
    \includegraphics[width=\linewidth]{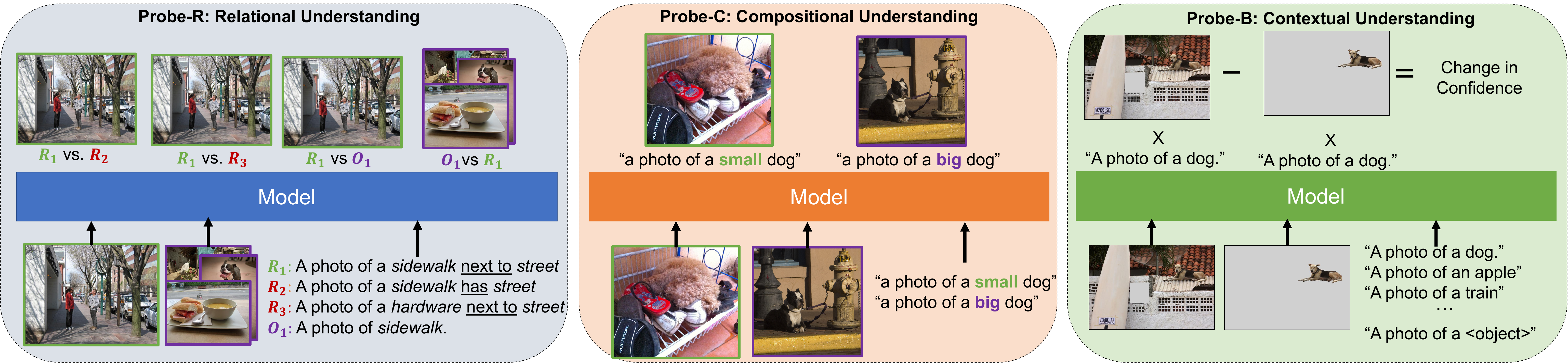}
    \caption{
   \textbf{ Overview of proposed benchmarks}. \textit{Probe-R} swaps the real subject or relation with an unlikely one and swaps a set of subject-only images to a subject-only prompt and the ground-truth relation prompt. \textit{Probe-C} asks the model to match two images and two prompts, swapping object or composition. \textit{Probe-B} compares object recognition performance before and after swapping out context from background and other surrounding objects.
    }
    \label{fig:dataset_overview}
\end{figure*}

\begin{table}[t!]
    \centering
    \caption{\textbf{Comparison of ours with various recent works} probing relational, attribute, and context understanding of models.}
    \label{tab:comparison}
\resizebox{\linewidth}{!}{\begin{tabular}{l|ccc}
\hline
\textbf{Approach} & \textbf{Relational} & \textbf{Compositional} & \textbf{Contextual} \\
\hline
{VL-CheckList \cite{vlchecklist}} & \checkmark & \checkmark & $\times$  \\
{ARO \cite{aro}} & \checkmark & \checkmark & $\times$ \\

{SVLC \cite{svlc}} & \checkmark & \checkmark & $\times$\\
{ControlledImCaps \cite{controlledimcaps}} & \checkmark & \checkmark & $\times$ \\
{CREPE \cite{crepe}} & \checkmark & \checkmark & $\times$ \\
{SugarCREPE \cite{sugarcrepe}} & \checkmark & \checkmark & $\times$ \\
\hline
{Ours} & \checkmark & \checkmark & \checkmark \\
\hline
\end{tabular}}
\end{table}

\section{Benchmark and Evaluation Metrics}
\label{datasets}
We evaluate three discrete concepts: object-relations, compositionality, and background context. We have generated three datasets: Probe-R, Probe-C, and Probe-B. A summary of each dataset is shown in Table \ref{tab:dataset_summary} and an overview in Figure \ref{fig:dataset_overview}. 

\textit{Prompting} 
is typically done in downstream image classification by forming sentences with each class name in the prompt,  such as ``a photo of a dog.''

The one with the highest similarity to the visual features is the predicted class \cite{clip,singh2022flava}.  
These benchmarks heavily rely on ``prompting'' the model by changing text input as well as image input in Probe-B.

\begin{table*}
    \centering
    \caption{\textbf{A summary of proposed benchmark datasets.} \textit{Tasks} include Image-Text matching (ITM), multi-label object recognition (MLR), and object recognition (R). \textit{Groups} refer to the group of images/text for each comparison being made. 
    Under \textit{attributes}, we list the dataset properties, where fillers are the types of replacements we use when removing background pixels.
    }
    \resizebox{\textwidth}{!}{\begin{tabular}
    {c|l|l|l|l|l|l|l}
    \hline
         \textbf{Dataset} & \textbf{Task} & \textbf{Description} & \textbf{Source} & \textbf{Images} & \textbf{Group Description} &\textbf{Groups} & \textbf{Attributes}    \\
         \hline
         Probe-R & ITM & Predicate/ Object Swapping & Visual Genome & 99,960 & 1 image, 10 pos. images, 4 prompts & 99,960 & 2,456 Relations, 6,006 Objects \\
         \multirow{2}{*}{Probe-C} & \multirow{2}{*}{ITM} & Composition Swapping & \multirow{2}{*}{MS COCO} & 40,681 & \multirow{2}{*}{2 images, 2 prompts} & 79,925 & \multirow{2}{*}{114 Compositions, 2,462 Objects} \\
         & & Object Swapping &  & 59,205 & & 375,607 & \\
         \multirow{2}{*}{Probe-B} & MLR & Background Removal & \multirow{2}{*}{MS COCO} & 31,745 & 3 images, 80 prompts & 31,745 & 4 fillers, 80 objects \\
          & R & Background+Object Removal &  & 1,484 & 3 images, $<$80 prompts & 9,375 & 4 fillers, 76 objects\\
          \hline
    \end{tabular}}
    \label{tab:dataset_summary}
\end{table*}

\subsection{Dataset}
\textbf{Probe-R: Relational Understanding}
\label{relational_dataset_overview}

To generate a dataset that can be used to probe for relational understanding, we collected samples from the Visual Genome \cite{visualgenome} dataset. These samples are used to probe whether models have learned consistent concepts of objects and their potential relationships to each other. 

For each group, we have four prompts $P \in \{R_1, R_2, R_3, O_1 \}$, one anchor image $X_{R_1}$ and 10 images $\mathbf{X}_{O_1}$ with the subject present and no other objects found in the anchor image. For each $X_{R_1}$, the ground truth relation $R_1=\langle s_1, r_1, o_1\rangle$ is compared to a swap of subject $R_3=\langle \overline{s}_1, r_1, o_1\rangle$ or predicate $R_2=\langle s_1, \overline{r}_1, o_1\rangle$. 
We sample $\overline{s}_1$ uniformly from subjects that do not occur in the dataset with $r_1$ and $o_1$ and similarly $\overline{r}_1$ is sampled uniformly from relations that do not occur with $s_1$ and $o_1$. This swapping of unlikely subjects and predicates allows us to test whether V+L models have learned consistent conceptual models of what object relations are possible in a system by comparing existing ones to unlikely ones. 
The final comparison is subject-only images $X_{O_1}$ to $P_{R_1}$ and a prompt with only the subject $P_{O_1}$. 

\noindent
\textbf{Probe-C: Compositional Understanding}
To generate a dataset that can be used to probe for compositional understanding, we collected samples from the MS COCO Captions dataset \cite{coco}. These samples are used to probe whether models have learned an understanding of object attributes and their relationships to each other. 
For each group, we have two images $x_1$ and $x_2$ and two prompts $p_1$ and $p_2$. This dataset has two splits, one where the compositions are swapped in the prompts and the other where objects are swapped. 
When swapping compositions, antonyms were manually mapped to each attribute to ensure that the attribute is not present in the image. For example, if there is a ``small dog'' in an image, the comparison could be ``a large dog.'' When swapping objects, the images must have the same composition but different objects. 

\noindent
\textbf{Probe-B: Contextual Understanding}
To generate a dataset that probes for model understanding on objects and their relationship to contextual cues found in an image's background, we collected samples from MS COCO \cite{coco} consisting of 80 objects. These samples are used to probe model reliance on background cues and reliance on co-occurrence between objects. 

For each group, there is an unmodified image $x_0$, an image with a random patch on the background $\tilde{x_0}$, a modified image where the background is removed $\tilde{x_1}$ and 80 or fewer prompts. We have two splits in this data, the first removing the background but keeping all objects Probe-B$_{MR}$ and the other removing both background and all other objects Probe-B$_R$. Probe-B$_R$ aims to probe models on whether they use conceptual maps on object co-occurrence to improve recognition. Probe-B$_{MR}$ aims to probe models on whether they have conceptual maps related to what group of objects are likely to be in what scenery or possible physical relations to each other. Poor performance on these tasks would indicate model use of such conceptual mappings, while good performance means they are focusing on object recognition only.

We experiment with four fillers: black, gray, gaussian noise, or a random scene. Random scenery was collected from the Indoor Scenes Dataset \cite{quattoni2009recognizing} and the Kaggle Landscape dataset \cite{arnaud58}. These images were manually filtered to ensure none of the 80 MS COCO classes were present. For single objects, images were only kept if the size of the object was between a threshold where the object was not too large and not too small relative to the image size. 

\subsection{Evaluation Metrics}
\label{metrics}
We use different evaluation metrics for each of the three datasets, but with a focus on \textit{change in model confidence}. This allows us to relate to the psychological paradigm ``violation-of-expectation'' (VoE) \cite{piloto2022intuitive,baillargeon1987object}. 
If V+Ls are learning conceptual models, then a violation of those models should be easily recognized and confidence should remain high when choosing between the correct prompt and the prompt that is violating expectation. For Probe-R, by probing models with data that is intended to violate expectation, we expect the confidence to remain high. For Probe-C, by probing models with paired opposites, we also expect the confidence to be high. For Probe-B, by removing visual information or replacing it with a violation of the original information, we expect the model to become confused and therefore the confidence to be low. 

\noindent
\textbf{Probe-R: }
We evaluate Probe-R using the mean confidence $\mu(c)$ and mean accuracy (acc) over all groups using equation \ref{eq:confidence_metric}, \ref{eq:new0}, and \ref{eq:new1}.
We compare one image $x$ to two prompts $p_1$ and $p_2$. 
\begin{align}
\footnotesize
\label{eq:confidence_metric}
    \left(c_1, c_2\right) &= \sigma(f(x, p_1), f(x, p_2)) \\
    \label{eq:new0}
    \mu(c) &=  \frac{1}{N}\sum_{i=1}^{N} c^i_1 \\
    \label{eq:new1}
    \text{acc} &=  \left\{ 
  \begin{array}{ c l }
    1 & \quad \textrm{if }  c_1 > c_2 \\
    0                 & \quad \textrm{otherwise}
  \end{array}\right. 
\end{align}
Logit scores from model $f$ are converted to softmax $\sigma$ predictions to measure the confidence $c_i$ of prompt $p_i$. Here $N$ denotes total number of images.

\noindent
\textbf{Probe-C: } To measure image and text matching between two images, $x_1$ and $x_2$, and two prompts, $p_1$ and $p_2$, using logit output from a model $f$, we adopt metrics from \cite{thrush2022winoground} measuring a text score ($t$), an image score ($i$), and a group score ($g$). $t$ measures the accuracy of the model selecting the correct prompt for a given image by equation \ref{eq:text_score}, \ref{eq:image_score}, and \ref{eq:group_score}.
\begin{align}
\footnotesize
\label{eq:text_score}
    t(p_1, x_1, p_2, x_2) &=  
    \left\{ 
  \begin{array}{ c l }
    1 & \quad \textrm{if } f(p_1, x_1) > f(p_2, x_1) \\ & \quad \text{ and } f(p_2, x_2) > f(p_1, x_2), \\
    0                 & \quad \textrm{otherwise}
  \end{array}
\right. \\
    \label{eq:image_score}
    i(p_1, x_1, p_2, x_2) &= 
    \left\{ 
  \begin{array}{ c l }
    1 & \quad \textrm{if } f(p_1, x_1) > f(p_1, x_2) \\ & \quad \text{ and } f(p_2, x_2) > f(p_2, x_1), \\
    0                 & \quad \textrm{otherwise}
  \end{array}
\right. \\  
    \label{eq:group_score}
    g(p_1, x_1, p_2, x_2) &= 
    \left\{ 
  \begin{array}{ c l }
    1 & \quad \textrm{if } t(p_1, x_1, p_2, x_2) \\ & \quad \text{ and } i(p_1, x_1, p_2, x_2), \\
    0                 & \quad \textrm{otherwise}
  \end{array}
\right.
\end{align}

\noindent
\textbf{Probe-B: }
We evaluate model reliance on either the co-occurrence of objects or background cues. Both tasks compare to both an original image $x_0$ and the original image with an added patch of the respective filler $\tilde{x_0}$ to take into account general robustness. $\tilde{x_1}$ will have either the background removed and replaced with a filler or have the background and all other objects replaced. The fillers are one of: ``black,'' ``gray,'' ``noise,'' or a random ``scene'' that does not have objects. 
The metrics we use for comparisons are the mean average precision (mAP) for multi-object recognition precision, relative robustness $\gamma^r$ measuring the relative drop/increase in performance (equation \ref{eq:set1_context_2}, and mean change in mAP $\mu(\triangle(c))$ (equation \ref{eq:set1_context_1}) for the objects. $\gamma^r$ and \textit{mAP} evaluates how much the models rely on background context to accurately describe the scenario. We collect the similarity between the image $x_n$ and for each object $o$ placed in a prompt $p_o \in \mathbf{p}$. This results in a set of similarity scores for each object prompt which is used to calculate the score of model's change in confidence $\triangle c$.

\begin{align}
\footnotesize
\label{eq:set1_context_2}
    \gamma^r &= 1 - \frac{h(x, \mathbf{p}) - h(\tilde{x}, \mathbf{p})}{h(x,  \mathbf{p})}  \\
    \label{eq:set1_context_1}
    \triangle \text{c}(x, \tilde{x}) &= \frac{1}{o} \sum_{i=1}^o  f(x, p_o) - f(\tilde{x}, p_o) 
\end{align}

\section{Benchmark Results}
Here we go through the models we are evaluating in this benchmark and then report the results of those models on the proposed datasets Probe-R, Probe-C, and Probe-B. 

\noindent
\textbf{Models}
\label{model_details}
We perform our experiments on ten recently developed and publicly available models: CLIP \cite{clip}, FLAVA \cite{singh2022flava}, ViLT \cite{kim2021vilt},  BridgeTower \cite{xu2022bridge}, BLIP \cite{blip}, BLIP2 \cite{blip2}, OTTER \cite{otter}, ALIGN \cite{align}, MetaCLIP \cite{metaclip}, and SigLIP \cite{siglip}. 

\textbf{CLIP}
\cite{clip} is a dual-stream, modality specific model that has a visual and text encoder of equal length and limited modality interaction. It 
uses a contrastive loss between text-image pairs as its only multimodal signal. 

\textbf{FLAVA} \cite{singh2022flava} is also a dual-stream encoder with an additional multimodal encoder that takes the ViT based \cite{dosovitskiy2020vit} single-stream encoders, merges them, and co-attends. It performs unimodal training for single-stream encoders followed by multimodal training on a global contrastive loss, a masked multimodal modeling task (MMM), and an image-text matching (ITM) loss. 

\textbf{ViLT} \cite{kim2021vilt} is a single-stream transformer that uses co-attention between modalities. It concatenates word embeddings and linear projections of image patches as input to a pre-trained ViT 
\cite{dosovitskiy2020vit}. It trains using an ITM loss, a masked language modeling (MLM) loss, and a word-patch alignment loss.

\textbf{Bridgetower} \cite{xu2022bridge} uses a dual-stream encoder with a multimodal encoder that incorporates the single-stream encoders at multiple layers using cross-attention based ``bridge layers.''
It uses a pre-trained ViT from CLIP as visual encoder, RoBERTa \cite{liu2019roberta} as text encoder, and is trained with MLM and ITM losses. 

\textbf{BLIP} \cite{blip} utilizes a mixture of encoder-decoder, and can operate in three functionalities: unimodal encoder, image-grounded text encoder, and image-grounded text decoder.

\textbf{BLIP2} \cite{blip2} uses a querying transformer that's at first trained in vision-language representation learning stage then vision-to language generative learning stage. It is a trainable module bridging the gap between the frozen image encoder and LLM.

\textbf{OTTER} \cite{otter} improves upon CLIP by using online entropic optimal transport to efficiently learn image-text pairs.

\textbf{ALIGN} \cite{align} is a dual-encoder which uses EfficientNet as image encoder and BERT as text encoder trained on a noisy dataset over one-billion image-text pairs.

\textbf{MetaCLIP} \cite{metaclip} follows CLIP by constructing metadata and carefully curating image-text pairs to imitate their dataset and training procedure.

\textbf{SigLIP} \cite{siglip} improves upon CLIP by introducing a pairwise Sigmoid loss instead of standard contrastive learning.

\subsection{Relational Evaluation}
\begin{figure*}
\begin{minipage}{0.5\textwidth}
\includegraphics[width=0.7\linewidth]{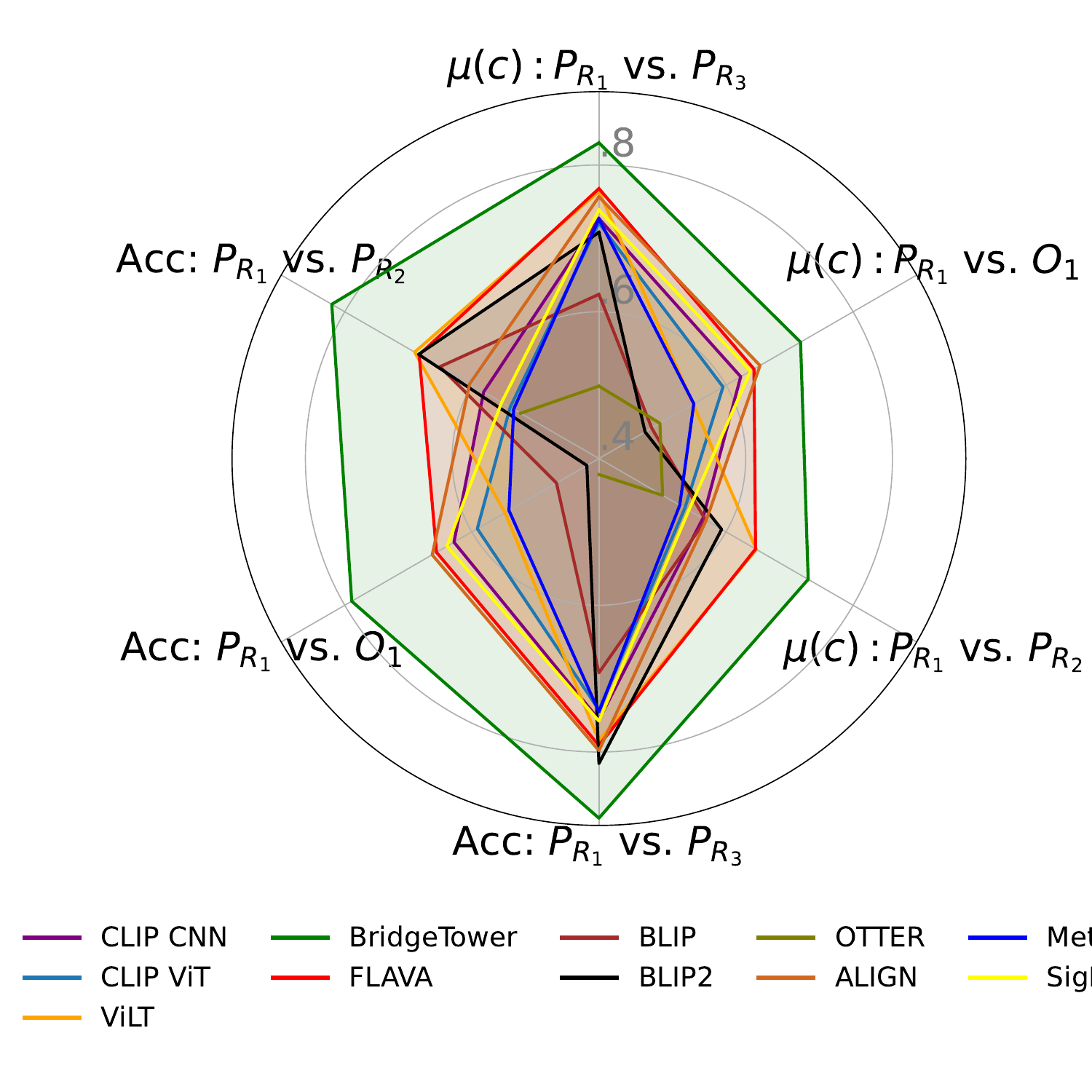}
\end{minipage}
\hfill
\begin{minipage}{0.4\textwidth}
     \centering
    \includegraphics[width={0.9\textwidth}]{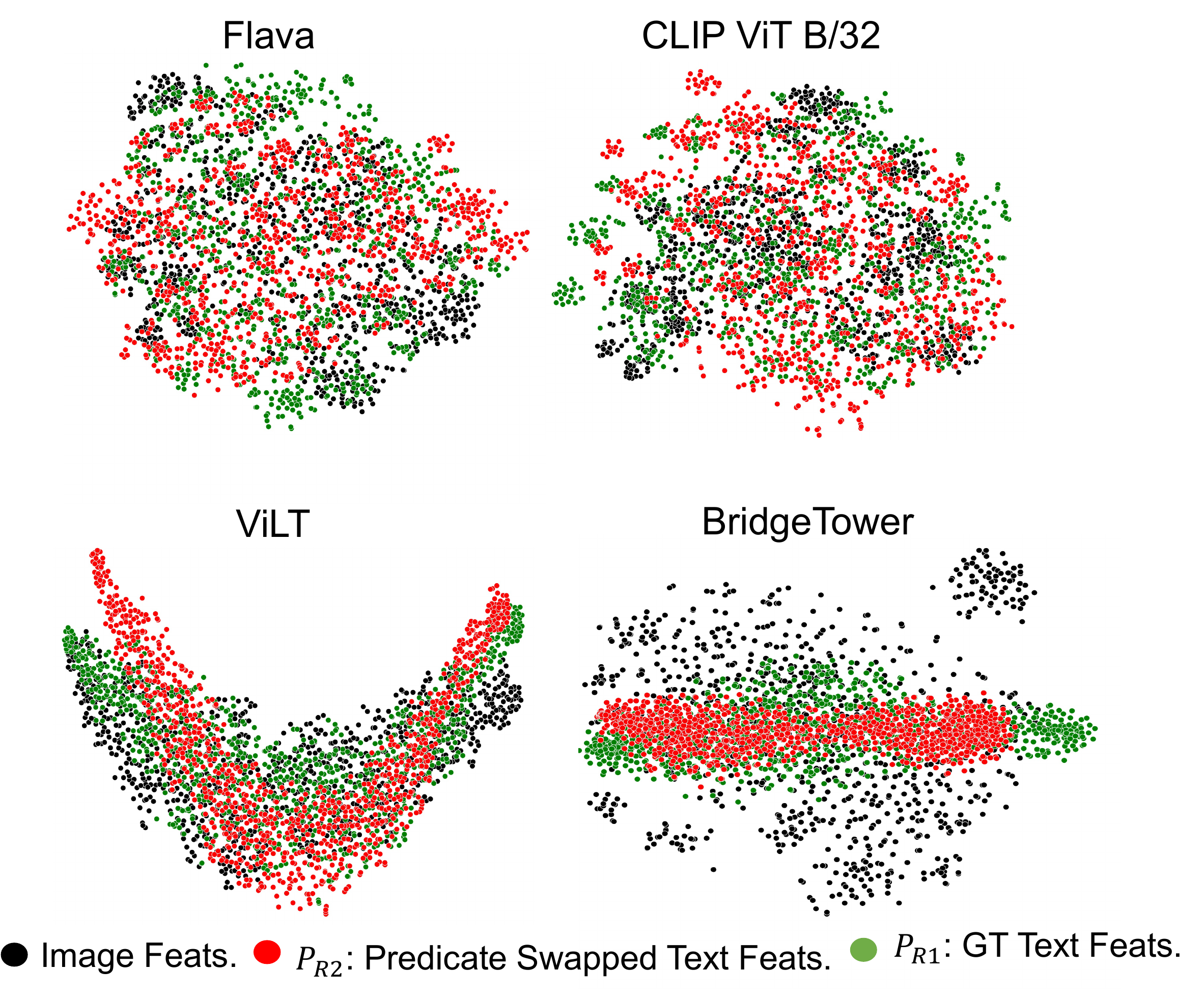}
\end{minipage}
    \caption{\textbf{Model's performance on relational understanding on Probe-R}. (left) Radar plot showing \textit{accuracy} and \textit{mean confidence $\mu(c)$} of different models. Here, the anchor image  $X_{R_1}$ contains the relation $R_1=\langle s, r, o \rangle$, image $X_{O_1}$ contains $O_1 = \langle s \rangle$. Prompts contain either the relation $P_{R_1}$, $P_{R_2}=\langle s, \overline{r}, o \rangle$, $P_{R_3}=\langle \overline{s}, r, o \rangle$, or $P_{O_1}=\langle s \rangle$. 
    (right) TSNE plot of the feature space for image features for some models where the prompt with the predicate swapped is denoted by $P_{R_2}$ and the ground truth prompt denoted by $P_{R_1}$. 
    }
    \label{fig:probe_r}
\end{figure*}

\noindent
\textbf{Models become confused when predicate is swapped, but more confident when object is swapped:} 
The overall results for the relation evaluation benchmark are shown in Figure \ref{fig:probe_r} (left) where it shows each model's accuracy and mean confidence $\mu(c)$ for matching the prompt to the anchor image $X_{R_1}$. When comparing an image to a correct prompt and an incorrect prompt where the relation/predicate is swapped with one not likely nor present in the image, the model's $\mu(c)$ for the correct prompt compared to incorrect is very low. This may indicated that selected models become ``confused'' when the relation is switched, even if it is a highly unlikely relation to even exist between the two objects. When swapping objects, the object that is swapped $\overline{s}$ is one that is highly unlikely, making this task simple if the model has a consistent understanding of what relationships are possible. Model confidence is higher when the object is swapped versus when the predicate is swapped. This may indicate that models are less confused when the task is specific to object recognition, focusing more on objects rather than the relationships between them. This may additionally indicate they are not understanding prompts as a ``whole'' but rather parts to a whole. 
To visualize the differences between models, we plot some of their feature space in Figure \ref{fig:probe_r} (right).
We see very different structures for BridgeTower and ViLT which heavily rely on cross-attention and image-text matching (ITM) when compared to FLAVA and CLIP. 

\noindent
\textbf{Summary: }BridgeTower and ViLT's performance indicates that co-attention is a method that can improve relational understanding. 
(1) This would indicate that both modality specific attention and co-attention simultaneously improves relational understanding.
(2) When the predicate is swapped to something that violates expectation, the drop in confidence, regardless of accuracy, indicates that their performance may not be due to an underlying conceptual map. (3) When the subject is swapped, all models show better performance compared to predicate swapping, indicating they are focusing on objects less-so than their relations to each other.

\subsection{Compositional Evaluation}
\textbf{Modality-specific attention and co-attention simultaneously greatly improves attribute-object relation understanding:} Overall results for evaluating model understanding of composition-object relationships are shown in Figure  \ref{fig:probe_c} (left) with additional results in the Supplementary. We show the image, group, and object scores for when the object (Obj.) is switched and for when the composition (Comp.) is switched. When presented with two images and two captions where the composition is the same but the objects are different, all models other than BridgeTower and BLIP2 perform on average double the performance versus when the composition is switched. This discrepancy indicates typically models are \textit{relying more on object recognition} when compositions are involved. BridgeTower and BLIP2's high performance indicates further support that a combination of modality-specific attention and cross-attention in parallel improves the learning of underlying concepts. 

\noindent
\textbf{Models stronger with more physical attributes like ``materials'' compared to visual-related like ``color'':} To better understand model failures when keeping the objects the same but swapping an attribute with its antonym, we categorized each attribute into 11 categories with results shown in Figure \ref{fig:probe_c} (middle). Attribute details are presented in the supplementary. 
All models struggle with ``visibility'' related compositions. The best performance was within the ``material'' and ``pattern'' category.

\noindent
\textbf{Transformers and CNNs differ on which attributes they understand best:}
To compare backbone models, we average the image scores over CLIP backbone architectures in Figure \ref{fig:probe_c} (right). Some noticeable patterns are that the CNN backbone models are better with ``material,'' ``pattern,'' and ``texture'' related compositions while ViT's are better at ``color'' and ``shape.'' This finding aligns with the findings of \cite{geirhos2018imagenet} where they found ImageNet trained CNNs are biased towards texture. 

\noindent
\textbf{Summary:} (1) Models struggle with compositionality but are better with those most associated with objects such as ``materials.'' (2) CNN based backbones may be better at recognizing texture and patterns while ViT backbones with color and shape. Surprisingly, (3) these models are typically better at matching captions given the image rather than text. 
\begin{figure*}
\centering
\begin{minipage}[t]{0.33\linewidth}
    \centering
    \includegraphics[width=\columnwidth]{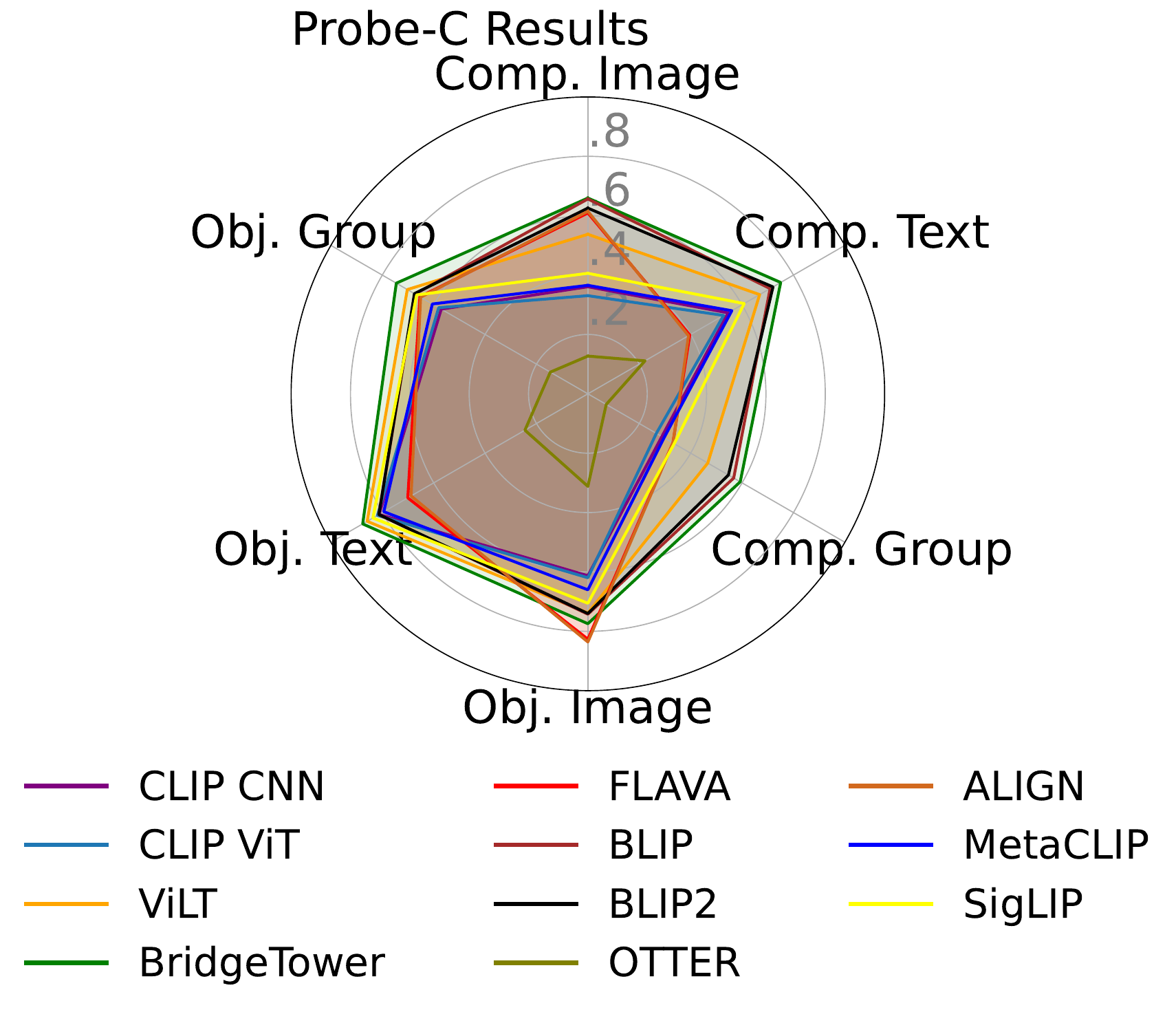}
\end{minipage}
\begin{minipage}[b]{.41\linewidth}
    \centering
    \includegraphics[width={\columnwidth}]{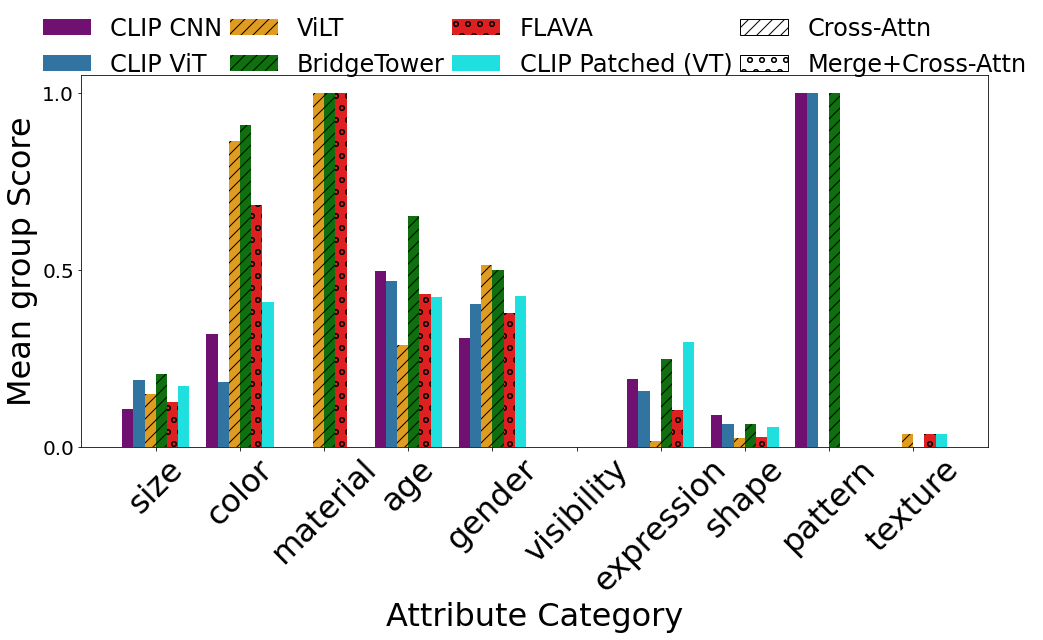}

\end{minipage}
\begin{minipage}[b]{.25\linewidth}
    \includegraphics[width=\columnwidth]{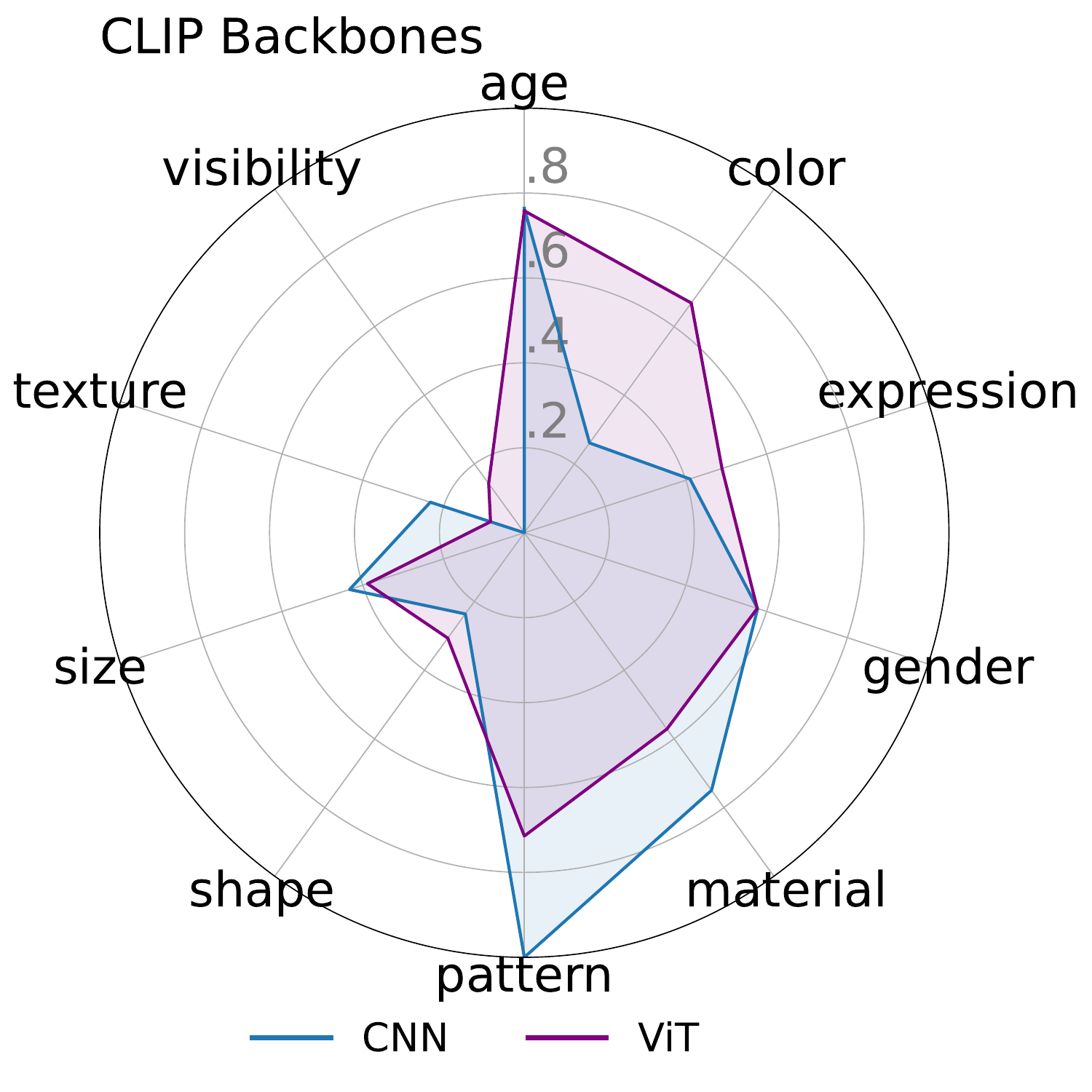}
\end{minipage}
\caption{\textbf{Model's performance on compositional understanding on Probe-C.} (left) The overall results for Probe-C showing the image, text, and group scores for when the object is swapped (\textit{Obj.}) or when the composition is swapped (\textit{Comp.}). (middle) Mean group score averaged across attribute categories. (right) 
CLIP scores averaged over different backbones. }
    \label{fig:probe_c}
\end{figure*}
\subsection{Background Context Evaluation}
\begin{figure*}
\centering
\hfill
\centering
\begin{minipage}{0.49\textwidth}

\centering
    \includegraphics[width=.7\linewidth]{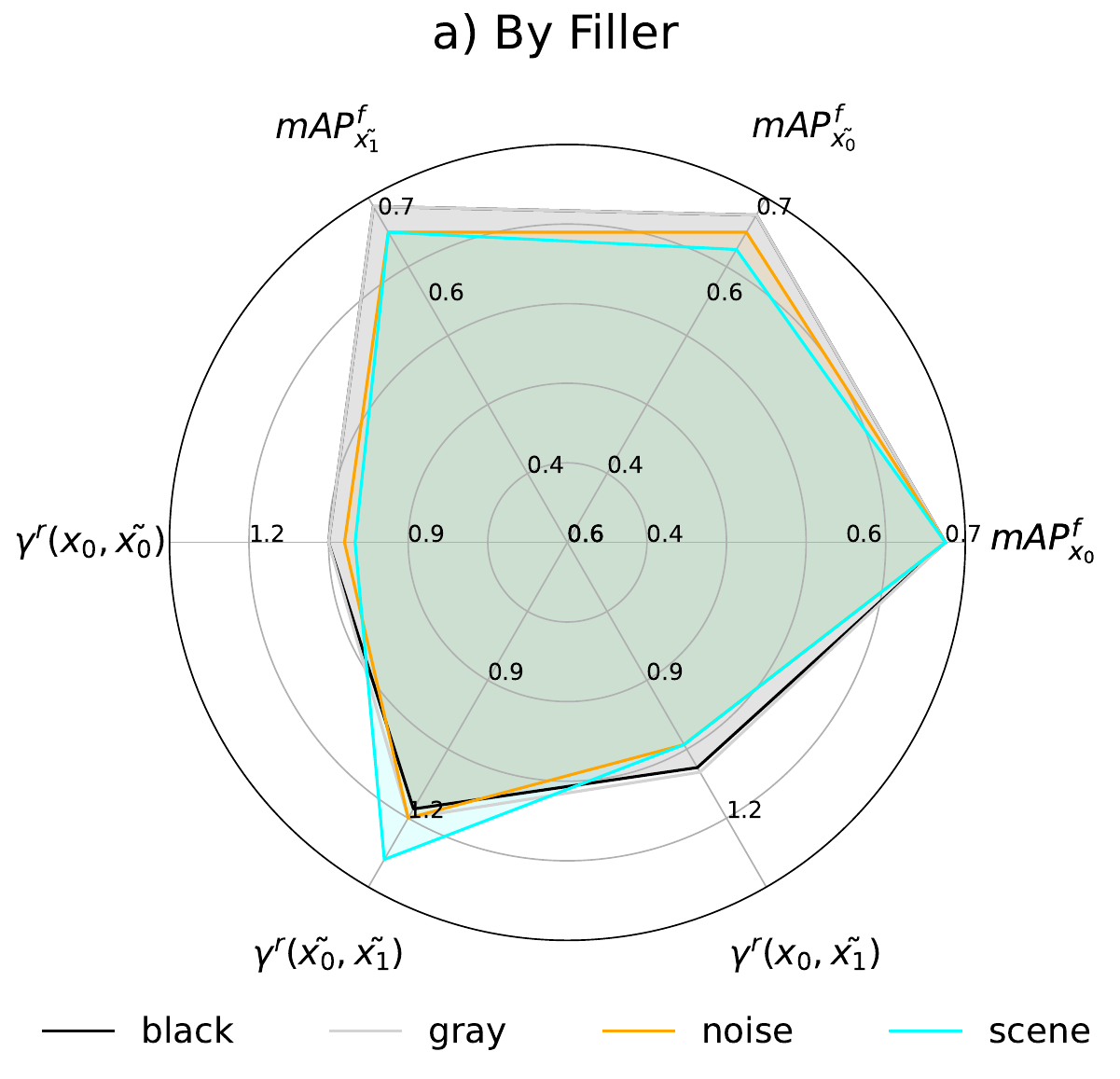}
\end{minipage}
\begin{minipage}{0.49\textwidth}

\centering
    \includegraphics[width=.8\linewidth]{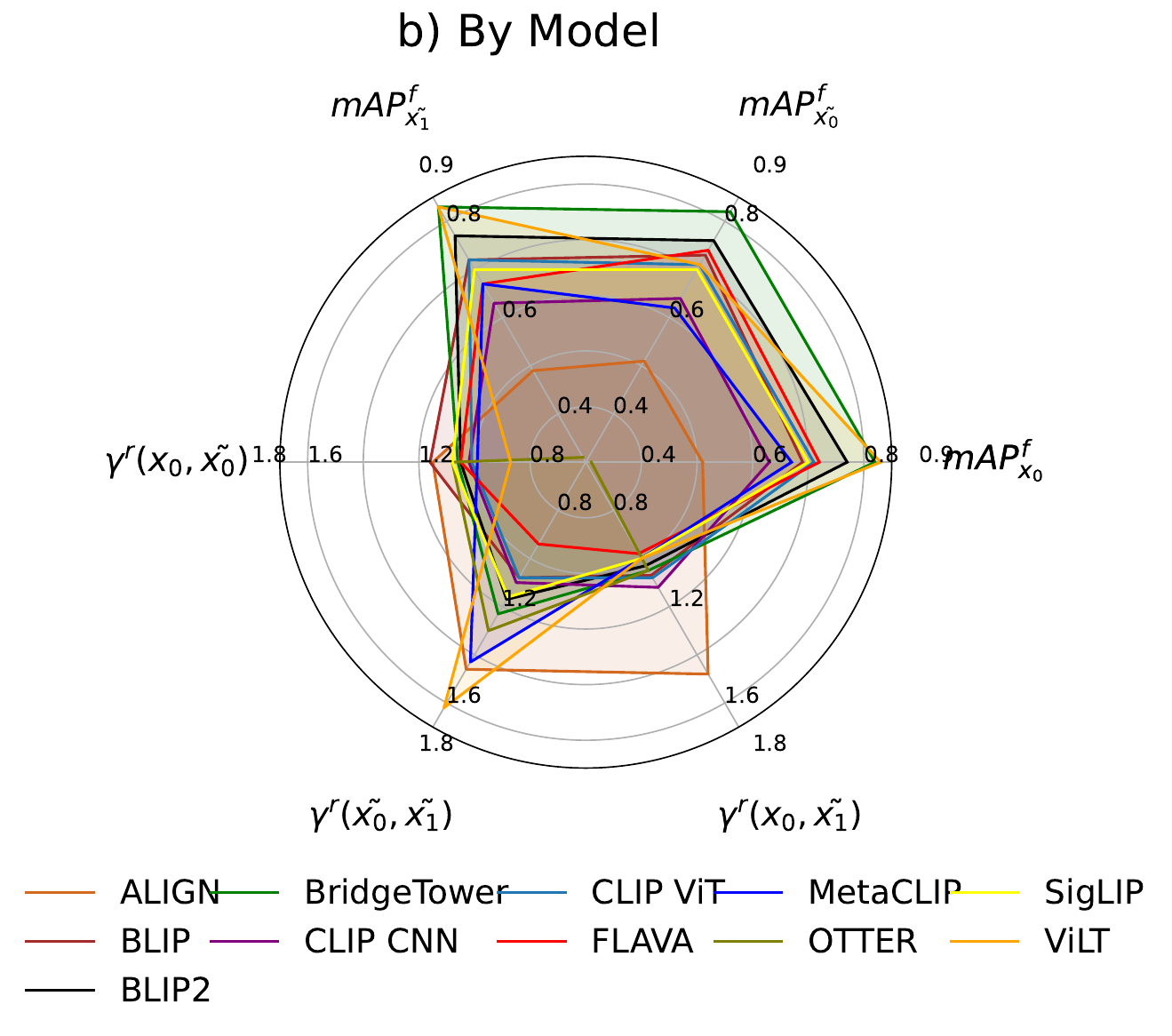}
\end{minipage}

\caption{\textbf{Model's performance on contextual understanding on Probe-B for only background removal.} (left) Mean results for replacing background with filler  and (right) for each model averaged over fillers. Comparisons between the original $x_0$, original+random patch $\tilde{x_0}$ and modified $\tilde{x_1}$. The metrics are \textit{mAP} and $\gamma^r$. 
}
\label{fig:filler_context}

\end{figure*}
\begin{figure*}
    \centering
    \begin{minipage}[b]{0.49\linewidth}
     \centering
    \includegraphics[width=.6\linewidth]{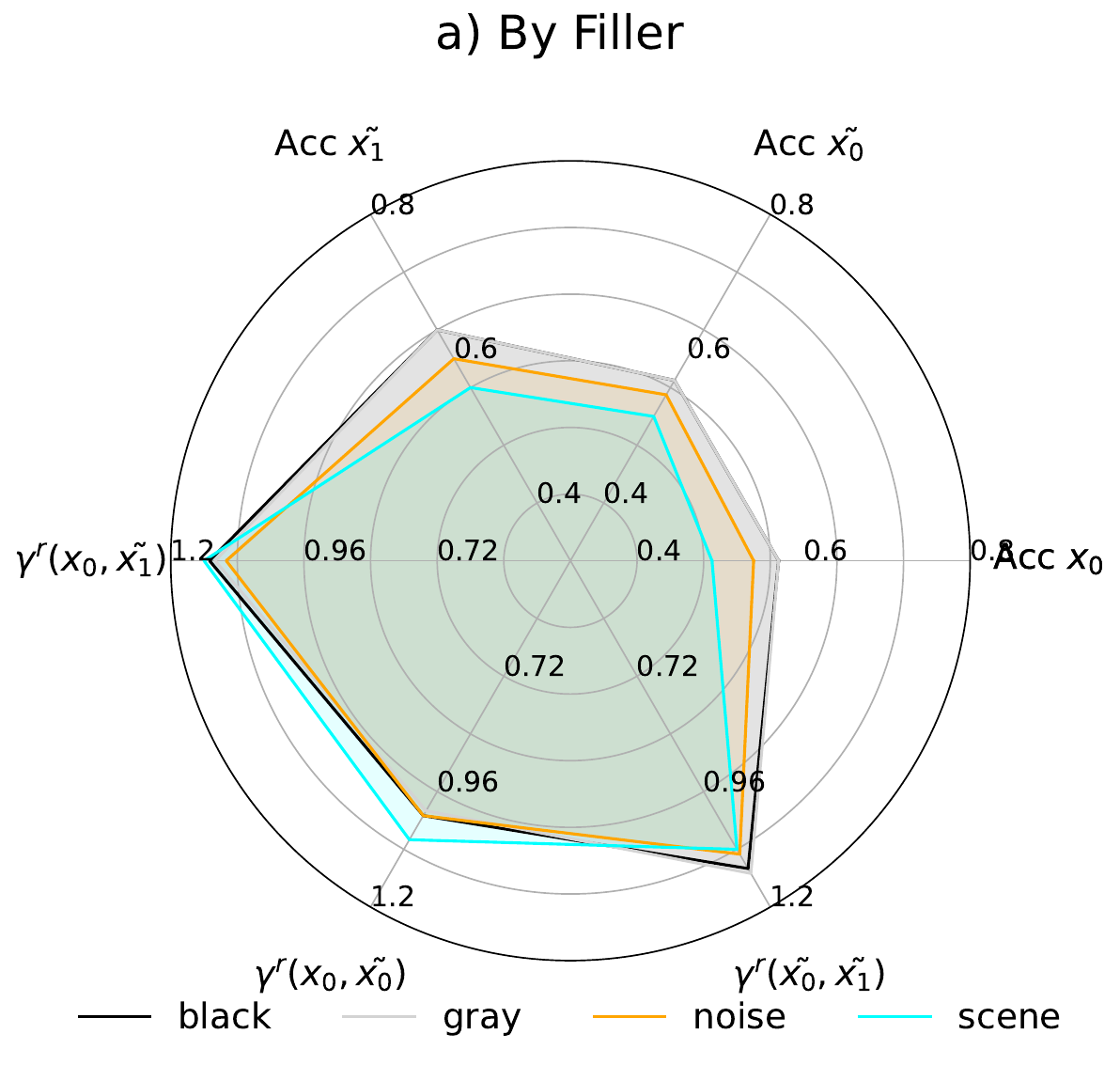}   
    \end{minipage}
    \begin{minipage}[b]{0.49\linewidth}
    \centering
    \includegraphics[width=0.8\linewidth]{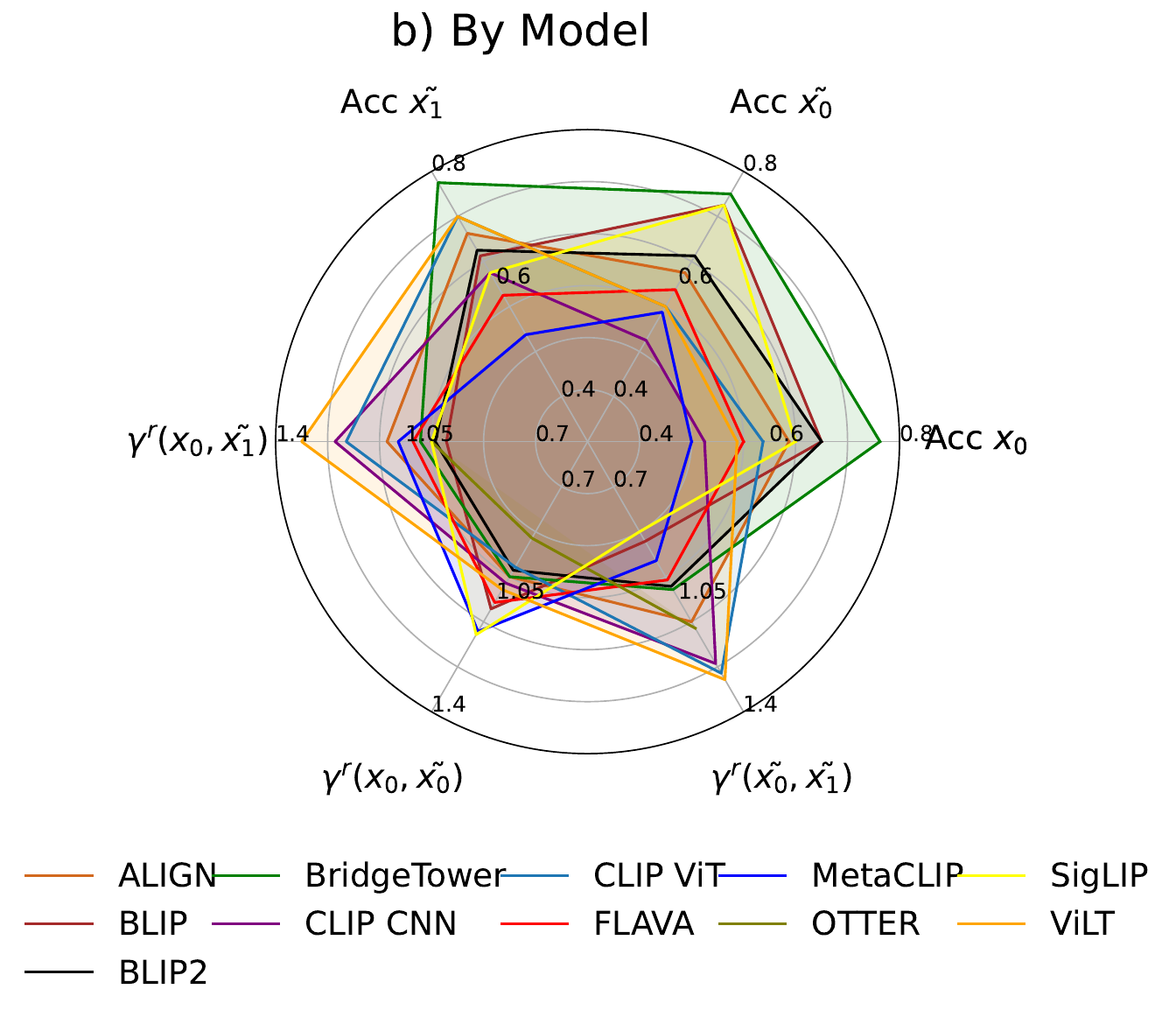}
    \end{minipage}
   
    \caption{\textbf{Model's performance on contextual understanding on Probe-B on background and all but one object removal.}(Left): Results for when the background and all other objects are replaced with a filler $\tilde{x_1}$, compared to the original $x_0$, and  (right) original+random patch $\tilde{x_0}$.} 

    \label{fig:coocccurence_all}
\end{figure*}

\begin{figure*}[t!]
    \centering
    \includegraphics[width=\linewidth]{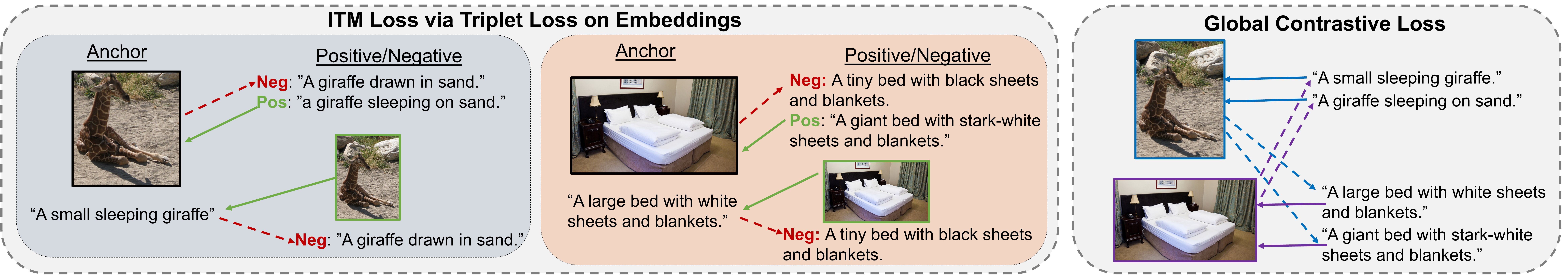}
    \caption{\textbf{Exploratory finetuning training scheme for CLIP.} Image-text matching (ITM) is used as a triplet loss whose pairings vary depending on if it is a compositional or a relational task. A contrastive loss is used to maintain general representations.
    }
    \label{fig:contrastive_training}

\end{figure*}
\begin{figure}
    \centering
\includegraphics[width=0.8\linewidth]{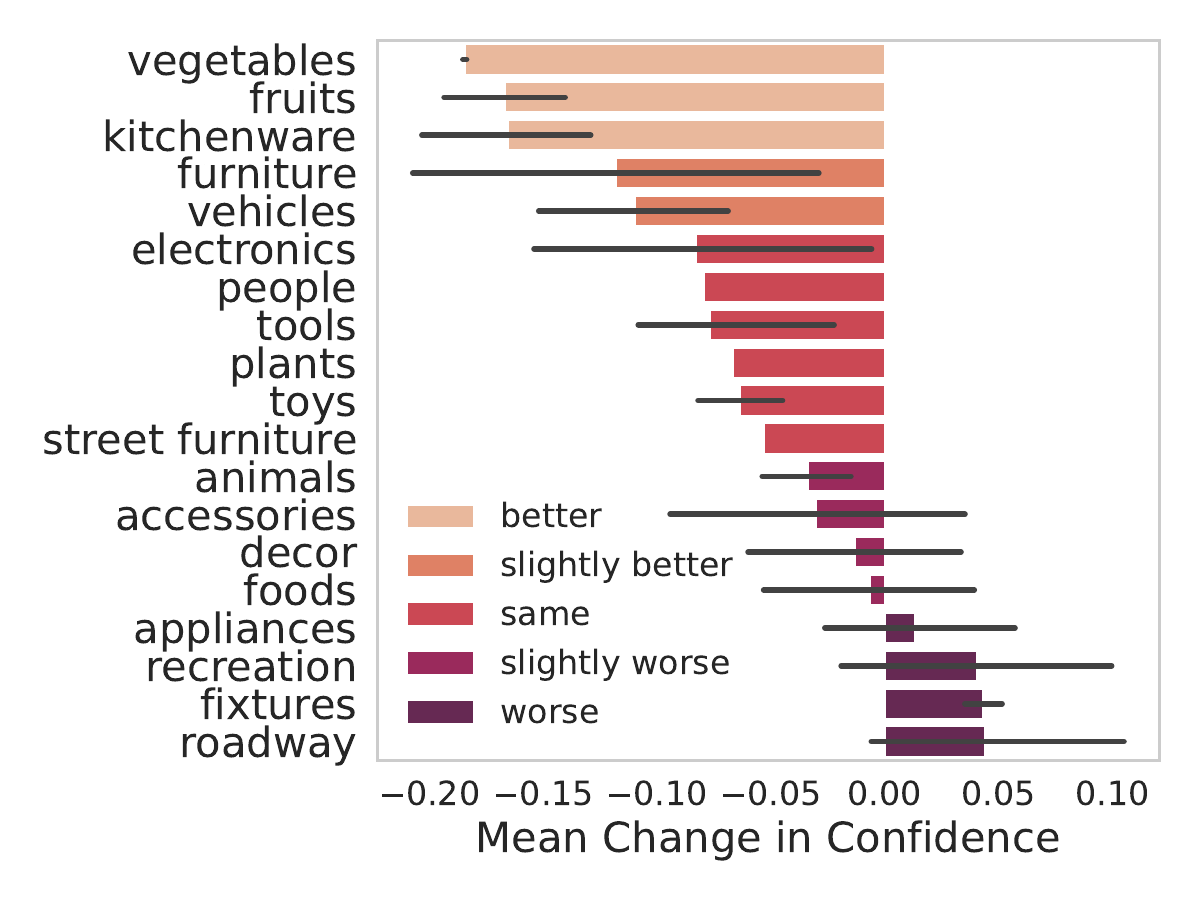}
    \caption{\textbf{Performance on contextual understanding for certain object categories. }Mean change in confidence ($\mu(\triangle c)$) from the ground truth $x_0$ to the modified $\tilde{x}_1$, where the background and other objects are removed.}
    \label{fig:probe-b-something}
\end{figure}

\noindent
\textbf{Models ignore what the background is replaced with, indicating little use of it:}
Overall results for evaluating model context understanding of background-object relationships are shown in Figure \ref{fig:filler_context} and \ref{fig:coocccurence_all}. 

Figure \ref{fig:filler_context} (left) shows the results averaged over filler type when only the background is removed. 
The most noticeable change is when comparing the ground truth image to $\tilde{x_0}$ and $\tilde{x_1}$ as expected. Overall, models are slightly less robust to when the background is replaced with either Gaussian noise or scenery. However, if models had underlying understanding of what objects belong in what context, models should be less robust to scenery. This indicates they may not have conceptual maps about objects and their relationship to context.

\noindent
\textbf{More co-attention may result in greater trade-off between robustness and performance:}
Figure \ref{fig:filler_context} (right) shows the overall results averaged over model type when only the background is removed. Similar to when looking at fillers, models are typically robust to background removal, indicating little use of context. ALIGN tends to improve when the background is removed. However, ViLT, ALIGN and MetaCLIP tend to be less robust when a patch is added to the image, noticeable even more so when the robustness between $\tilde{x_0}$ and $\tilde{x_1}$ is so high. This appears to be a trade-off between robustness and performance. 

\noindent
\textbf{Some objects benefit from the presence of others, but most are better off without:}
Figure \ref{fig:coocccurence_all} shows the overall results for when the background and all other objects but one are removed, averaged over either filler (left) or model (middle). When averaging over filler, models appear to be more robust when detecting one object as opposed to multiple objects in an image. When averaging scores over models, $\gamma^r$ tends to be over 1 when comparing to the background removed image $\tilde{x_1}$, indicating models improve when objects are in isolation. This case is especially prominent for ALIGN. This indicates that models may be distracted from background information rather than using it for object recognition. In order to better understand what objects models are using background with more than others, we categorize objects into sub-categories as shown in Figure \ref{fig:probe-b-something}. The object-types that models struggle with most appear to be large objects used in a common setting, such as ``ovens'' for appliances and ``sink'' for fixtures. This may indicate that there is some context used but for certain objects more than others. 

\noindent
\textbf{Summary}
(1) Models tend to not use context in order to recognize multiple objects but (2) for some individual objects, models do use context. (3) These models are typically robust to a change in background where models like ViLT, ALIGN, and BridgeTower are more susceptible to a particular patch being changed. (4) When objects are placed in random scenery that violates-expectation, models still perform similarly to when the original background is there. This may indicate that overall, models are not learning conceptual maps relating objects to their context. 

\section{Finetuning for better conceptual understanding}
Dual-stream encoders like CLIP and FLAVA allow  uni-modal feature representations that can be extracted and used for a variety of downstream tasks. Improving models that do not require paired input would provide greater value and stronger representations. To explore this idea, we finetune (FT) CLIP ViT-B/32 on a new dataset inspired by this benchmark called RelComp. The new dataset RelComp for attribute-object and object-object relations is based on MS COCO \cite{coco} and VisualGenome \cite{visualgenome} and has no overlap between the benchmark datasets.
\begin{table}
    \centering 
    \caption{\textbf{Performance on finetuned and patched CLIP on proposed RelComp dataset.} ImageNet accuracy is shown to measure the drift from the original CLIP space. RelComp and Probe-C/R respectively report image score and mean accuracy for the correct image-to-prompt matching.}
    \label{tab:training_results}
    \resizebox{\columnwidth}{!}{\begin{tabular}{lcccc}
        \hline
        Model & ImageNet & RelComp & Probe-C  & Probe-R  \\
        \hline
        ViLT & -- & 76.00 & 90.78 & 69.00 \\
        BridgeTower & -- & 85.00 & 90.06 & 82.20 \\
        \hline\hline
        FLAVA & 56.83 & 47.12 & 83.85 & 68.29 \\
        CLIP ViT B32 & \textbf{63.60} & 51.93 & 88.15 & 53.52 \\
        \hline
        CLIP Patched (T) & 57.85 & \textbf{67.85 }& 89.49 & \underline{71.14} \\
        CLIP Patched (V) & \underline{61.45} & 54.66 & \underline{89.81} &  61.40\\
        CLIP Patched (VT) & 54.61 & \underline{64.27} & \textbf{90.30} & \textbf{71.20} \\
        \hline
    \end{tabular}}
\end{table}
 
We propose using selective negative and positive pairing based on attribute and predicate swaps and finetune using image-text matching (ITM) loss and a contrastive loss (C) \cite{clip,singh2022flava} for finetuning (see Figure \ref{fig:contrastive_training}).  
We linearly interpolate the original CLIP weights with our FT weights using an alpha$=0.2$ to prevent ``catastrophic forgetting'' \cite{ilharco2022patching,wortsman2022robust}. We call this ``CLIP Patched'' and finetune visual-encoder only (V), text-encoder only (T), or both (VT). More details about losses, implementation, and dataset are in the Supplementary.

\noindent
Overall results for our exploratory experiment are shown in Table \ref{tab:training_results}. We observe drift as measured by ImageNet accuracy, even when patching. When finetuning using the visual-encoder only, the drift is less pronounced, but so is the improvement on RelComp. The largest increase is seen with FT text encoder only. This may indicate that for non-cross-attention models, text is more important for conceptual mapping. Our findings indicate that by using selective negative sampling we can enforce compositional and relational learning without extensive co-attention and computational complexity.

\section{Conclusions}
In this benchmark we evaluated large visual-language (V+L) models on relational, compositional, and contextual understanding with three new datasets: Probe-C, Probe-R, and Probe-B. For compositional understanding, we observe (1) models struggle with compositionality. (2) CNN backbones may be better at recognizing texture and patterns while ViT backbones are with color and shape. For relational understanding, we observe (1) both modality specific attention and co-attention in parallel improves relational understanding. (2) An expectation violating predicate swap surfaces the lack of a conceptual map through drop in confidence. For contextual understanding we observe (1) models mostly tend to not use context in order to recognize multiple objects. (2) When objects are placed in random scenery that violates expectation, model performance is unchanged, indicating a lack of conceptual map of context. 
When trying to improve CLIP, the dual-encoder with no cross-attention, by finetuning on our proposed selective negatives training paradigm on the proposed RelComp dataset, (1) we find that there is a small drop in classification performance, but (2) an improvement on Probe-R, Probe-C, and RelComp is observed, indicating an improvement in relational and compositional learning.
We hope these insights will help drive future work on building V+L models that better ``understand.''

{
    \small
    \bibliographystyle{ieeenat_fullname}
    \bibliography{main}
}

\clearpage
\setcounter{page}{1}
\maketitlesupplementary

The supplementary will provide additional details about our proposed datasets, finetuning CLIP and the models evaluated on in this benchmark. Additional details and results for Probe-R, Probe-C and Probe-B are in Section \ref{supp:datasets}. We provide more details about finetuning CLIP and additional results in Section \ref{supp:clip_patched}. In Section \ref{supp:model_details} we provide additional details about the models we evaluated in this benchmark. 

\section{Datasets Details}
\label{supp:datasets}
In this section we will provide additional results for the different dataset benchmarks. 

\subsection{Probe-R: Relational Understanding}
This dataset was created using Visual Genome (VG) \cite{visualgenome}. To collect unlikely ``$<$subject, predicate, object$>$'' triplets, we first cleaned the relationship aliases. This was done by mapping repeated aliases that meant the same thing into one, for example ``are standing next to'' would become ``standing next to''. This was done to reduce the space to map all objects to aliases they have been associated with as well as to confirm they have not been associated with one similar. We then collect all the objects each cleaned alias was associated with using regex and NLTK part-of-speech (POS) tagging \cite{bird2009natural}. Using these object collections, we iterated through $100,000$ VG annotations of $R_1=\langle s_1, r_1, o_1\rangle$ to (1) replace the existing alias with an alias that the current subject and object are not associated with as swap ($R_2=\langle s_1, \overline{r}_1, o_1\rangle$) and (2) replace the existing subject with an object that is not associated with the current alias ($R_3=\langle \overline{s}_1, r_1, o_1\rangle$). To better collect images with specific objects in them, we iterated through VG and generated a mapping of each image ID to all objects present in the image according to the relationships annotations. We extract positive images $X_{O_1}$ that do not have the relation but have the subject and no other objects present in the anchor image $X_{R_1}$.

\begin{figure*}
    \includegraphics[width=\textwidth]{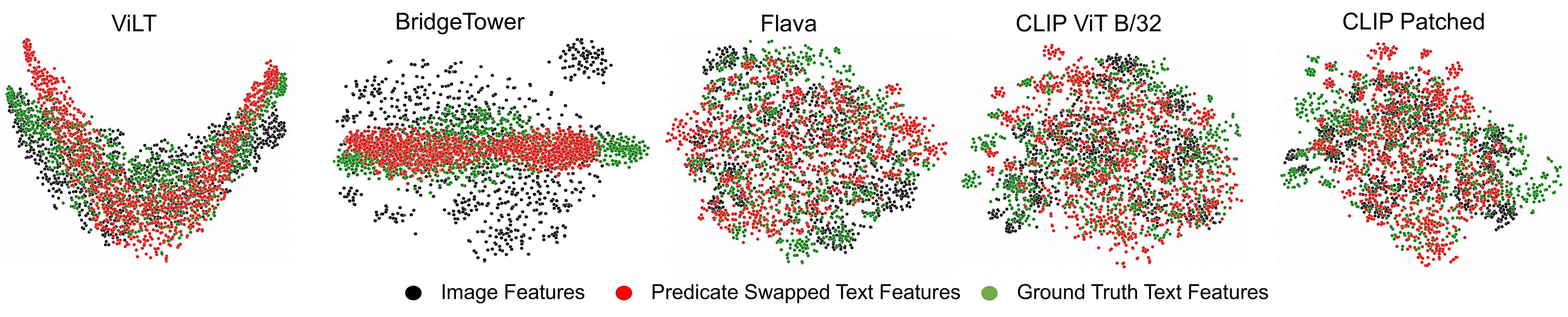}
    \captionof{figure}{TSNE plots of image features (black) and text features from the Probe-R. Text features are prompts generated from either the ground truth relation $R_1$ (green) or the relation with the predicate swapped to an unrealistic one $R_2$ (red). Both ViLT and BridgeTower rely on cross-attention heavily, showing the impact on the feature space. While the features for the other models are more visibly in the same space, ViLT and BridgeTower generally show higher performance. CLIP patched is finetuning both visual and text encoders using RelComp and patching with an alpha of $0.2$ \cite{ilharco2022patching}. CLIP ViT is ViT/L-14@336px while CLIP CNN is RN50x4.}
    \label{fig:relation_tsne}
\end{figure*}

\begin{table*}[t!]
    \centering
    \caption{\textbf{Overall results for relation evaluation.} The anchor image $X_{R_1}$ contains the relation $R_1=\langle s, r, o \rangle$, image $X_{O_1}$ contains $O_1 = \langle s \rangle$. Prompts contain either the relation $P_{R_1}$, $P_{R_2}=\langle s, \overline{r}, o \rangle$, $P_{R_3}=\langle \overline{s}, r, o \rangle$, $P_{O_1}=\langle s \rangle$, or $P_{O_3}=\langle \overline{s} \rangle$. The mean confidence $\mu(c)$ is for the correct prompt to image. Models with CLIP Patched are those we finetuned on our training dataset RelComp. We finetuned either the text encoder (T), the visual encoder (V) or both (VT). Models show higher performance for when objects are switched but lower performance when the relation is switched, showing the models are confused.
    }
    \resizebox{.85\textwidth}{!}{\begin{tabular}{l|ll|ll|ll|ll|}
 \multirow{3}{*}{Model} & \multicolumn{6}{c|}{$X_{R_1}$} & \multicolumn{2}{c|}{$X_{O_1}$}  \\
 
\cline{2-9}
  & \multicolumn{2}{c|}{$P_{R_1}$ vs. $P_{R_3}$} & \multicolumn{2}{c|}{$P_{R_1}$ vs. $P_{R_2}$}  & \multicolumn{2}{c|}{$P_{R_1}$ vs. $P_{O_1}$}  & \multicolumn{2}{c|}{$P_{R_1}$ vs. $P_{R_2}$} \\ 
\cline{2-9}
  & $\mu(c)$ & Acc & $\mu(c)$ & Acc & $\mu(c)$ & Acc  & $\mu(c)$ & Acc    \\

\hline
CLIP RN50         &                          $69.77$ &                     $72.14$ &                          $51.33$ &                     $51.13$ &                          $61.19$ &                     $61.69$ &                           $78.44$ &                      $89.10$ \\
CLIP ViT L/14     &                          $71.59$ &                     $73.68$ &                          $52.44$ &                     $52.59$ &                          $59.09$ &                     $58.67$ &                           $84.17$ &                      $93.23$ \\
CLIP ViT-B/16     &                          $71.08$ &                     $73.40$ &                          $52.84$ &                     $53.37$ &                          $61.69$ &                     $62.07$ &                           $79.62$ &                      $89.96$ \\
CLIP ViT/B-32     &                          $69.00$ &                     $71.21$ &                          $53.02$ &                     $53.53$ &                          $58.83$ &                     $58.56$ &                           $82.21$ &                      $92.32$ \\
CLIP ViT          &                          $72.09$ &                     $74.27$ &                          $53.52$ &                     $53.97$ &                          $59.53$ &                     $59.14$ &                           $83.97$ &                      $93.50$ \\
CLIP RN101        &                          $70.62$ &                     $73.28$ &                          $54.01$ &                     $55.11$ &                          $60.66$ &                     $60.83$ &                           $79.08$ &                      $91.17$ \\
CLIP RN50x64      &                          $72.79$ &                     $74.79$ &                          $56.66$ &                     $58.03$ &                          $64.10$ &                     $64.88$ &                           $78.10$ &                      $87.15$ \\
CLIP CNN          &                          $72.71$ &                     $75.59$ &                          $56.35$ &                     $58.14$ &                          $62.31$ &                     $62.81$ &                           $78.29$ &                      $90.77$ \\
CLIP RN50x16      &                          $73.91$ &                     $76.57$ &                          $58.08$ &                     $60.52$ &                          $59.80$ &                     $59.79$ &                           $83.03$ &                      $94.05$ \\
CLIP Patched (V)  &                          $78.58$ &                     $81.41$ &                          $59.36$ &                     $62.27$ &                          $66.56$ &                     $68.07$ &                           $81.32$ &                      $90.79$ \\
FLAVA             &                          $76.79$ &                     $79.09$ &                          $64.65$ &                     $68.29$ &                          $64.40$ &                     $65.56$ &               84.19 &                      $90.12$ \\
ViLT              &                          $76.41$ &                     $78.45$ &                          $64.77$ &                     $69.00$ &                          $54.84$ &                     $54.78$ &                  $\mathbf{94.23}$ &             $\mathbf{99.10}$ \\
CLIP Patched (VT) &                          $80.56$ &                     $84.46$ &                          $64.53$ &                     $71.12$ &                          $67.63$ &                     $70.07$ &                           $81.74$ &                      $92.40$ \\
CLIP Patched (T)  &              $\underline{82.37}$ &         $\underline{86.25}$ &              $\underline{66.28}$ &         $\underline{72.55}$ &              $\underline{67.93}$ &         $\underline{70.51}$ &                           $79.76$ &                      $90.70$ \\
BridgeTower       &                 $\mathbf{83.03}$ &            $\mathbf{89.01}$ &                 $\mathbf{72.93}$ &            $\mathbf{82.04}$ &                 $\mathbf{71.73}$ &            $\mathbf{78.90}$ &                           $76.58$ &          94.38 \\
BLIP   & $62.38$ & $69.2$  & $56.8$  & $65.02$ & $48.31$ & $46.68$ & $76.65$ & $\underline{97.15}$ \\
BLIP2    & $70.82$ & $81.57$ & $59.31$ & $68.39$ & $47.26$ & $41.9$  & $78.06$ & $96.51$ \\
OTTER   & $49.87$ & $42.18$ & $50.02$ & $52.33$ & $49.6$ & $24.48$ & $50.49$ & $84.28$ \\
ALIGN    & $75.68$ & $79.81$ & $56.88$ & $60.34$ & $65.35$ & $66.24$ & $73.42$ & $90.51$ \\
MetaCLIP & $72.66$ & $74.53$ & $52.72$ & $53.42$ & $54.93$ & $54.16$ & $\underline{88.68}$ & $96.14$ \\
SigLIP   & $73.88$ & $75.78$ & $54.14$ & $55.31$ & $63.86$ & $63.92$ & $82.77$ & $91.86$\\

\end{tabular}}

    \label{tab:sup_relation_eval}
\end{table*}

\begin{table*}[t!]
    \centering
    \caption{\textbf{Overall results for the compositional evaluation} on select models with highest scores in \textbf{bold} and second highest \underline{underlined}. Mean confidence for the correct prompt-to-image is $\mu\left( c \right)$.  CLIP ViT is ViT/L-14@336px while CLIP CNN is RN50x4.
}

\resizebox{0.9\textwidth}{!}{\begin{tabular}{l|rrrr|rrrr}

&  \multicolumn{4}{|c|}{Composition Switch} & \multicolumn{4}{c}{Object Switch} \\
  Model &  $\mu \left( c \right) \uparrow$ &  Image $\uparrow$ &  Text $\uparrow$ &  Group $\uparrow$ &  $\mu \left( c \right) \uparrow$ & Image $\uparrow$ &  Text $\uparrow$ &  Group $\uparrow$\\
\hline
CLIP ViT          &                     $69.69$ &              $33.06$ &              $52.82$ &              $26.64$ &                $88.15$ &              $61.96$ &              $81.89$ &              $58.05$ \\
CLIP RN50         &                     $69.47$ &              $33.41$ &              $54.60$ &              $26.92$ &                $87.00$ &              $61.40$ &              $80.17$ &              $56.81$ \\
CLIP ViT-B/16     &                     $69.23$ &              $34.29$ &              $52.23$ &              $26.94$ &                $88.02$ &              $63.53$ &              $81.44$ &              $59.12$ \\
CLIP ViT L/14     &                     $69.41$ &              $33.73$ &              $52.36$ &              $27.01$ &                $87.89$ &              $61.93$ &              $81.31$ &              $57.86$ \\
CLIP RN101        &                     $69.24$ &              $34.95$ &              $51.82$ &              $27.42$ &                $86.99$ &              $61.75$ &              $80.58$ &              $57.46$ \\
CLIP RN50x64      &                     $70.44$ &              $35.21$ &              $52.89$ &              $27.95$ &                $87.75$ &              $63.09$ &              $80.55$ &              $58.27$ \\
CLIP ViT/B-32     &                     $69.79$ &              $34.71$ &              $53.85$ &              $27.96$ &                $87.75$ &              $62.01$ &              $80.92$ &              $57.65$ \\
CLIP RN50x16      &                     $69.77$ &              $35.51$ &              $53.24$ &              $28.07$ &                $87.91$ &              $63.12$ &              $82.23$ &              $59.38$ \\
CLIP CNN          &                     $69.75$ &              $36.07$ &              $54.56$ &              $28.79$ &                $87.24$ &              $61.29$ &              $81.24$ &              $57.06$ \\
FLAVA             &                     $67.45$ &  $60.93$ &              $39.65$ &              $33.09$ &                $83.85$ &     $\underline{82.66}$ &              $70.08$ &              $65.37$ \\
CLIP Patched (T)  &                     $71.94$ &              $40.96$ &              $58.79$ &              $33.83$ &                $89.58$ &              $68.81$ &              $84.36$ &              $65.19$ \\
CLIP Patched (V)  &                     $73.65$ &              $42.30$ &              $59.10$ &              $34.48$ &                $89.79$ &              $66.17$ &              $84.00$ &              $62.45$ \\
CLIP Patched (VT) &                     $73.65$ &              $44.53$ &              $61.92$ &              $37.18$ &    $\underline{90.30}$ &              $70.01$ &              $85.41$ &              $66.83$ \\
ViLT              &         $\underline{79.02}$ &              $53.74$ &  $66.84$ &  $46.65$ &       $\mathbf{90.78}$ &              $73.82$ &  $\underline{85.88}$ &  $\underline{70.26}$ \\
BridgeTower       &            $\mathbf{81.88}$ &     $\mathbf{65.95}$ &     $\mathbf{75.02}$ &     $\mathbf{59.28}$ &                $90.05$ &  $77.44$ &     $\mathbf{87.63}$ &     $\mathbf{74.54}$ \\
BLIP     & $73.1$  & $\underline{65.64}$ & $70.91$ & $\underline{56.74}$ & $81.59$ & $74.24$ & $81.37$ & $47.26$ \\
BLIP2    & $70.98$ & $62.55$ & $\underline{72.03}$ & $54.69$ & $81.8$  & $74.01$ & $81.31$ & $67.5$  \\
OTTER    & $50.05$ & $12.71$ & $22.24$ & $7.14$  & $50.21$ & $31.17$ & $24.5$  & $14.62$ \\
ALIGN    & $71.85$ & $61.48$ & $39.16$ & $33.13$ & $87.9$  & $\mathbf{83.6}$ & $68.79$ & $65.08$ \\
MetaCLIP & $71.56$ & $36.55$ & $56.01$ & $29.63$ & $87.31$ & $66.02$ & $79.41$ & $60.53$ \\
SigLIP   & $74.5$  & $40.59$ & $60.82$ & $33.59$ & $90.1$  & $70.55$ & $83.65$ & $66.64$ \\

\end{tabular}}

    \label{tab:compositional_eval}
    \end{table*}

\begin{table*}[]
    \centering
    \caption{\textbf{The attributes that belong to each category for the compositional analysis} on specific attributes in Probe-C.}
    \resizebox{\linewidth}{!}{\begin{tabular}{l|p{20cm}|l}
Attribute & Category & Groups\\
\hline
            age &                                                                                                                                                                                                        [young, old, new] &   2,051 \\ \hline
     color & [greyscale, coloured, sepia, reddish, bronze, greenish, green, turquoise, blue, tan, red, white, silver, purple, gold, pink, navy, brown, teal, gray, black, yellow, grey, golden, camo, pinkish, beige, orange, blonde] &  39,971 \\ \hline
expression &                                                                                                                                                                         [happy, unhappy, smiling, laughing, smiley, sad] &   2,088 \\ \hline
    gender &                                                                                                                                                                                                           [male, female] &   2,346 \\ \hline
  material &                                                                                                  [tin, aluminum, cloth, gravel, unpaved, wooden, stainless, marble, metallic, metal, grassy, porcelain, wooded, pebbled] &   3,875 \\ \hline
   pattern &                                                                                                                                                  [checkered, patterned, striped, spotted, plaid, stripped, checkerboard] &    3,08 \\ \hline
     shape &                                                                                                                                         [triangular, flat, circular, triangle, oval, round, dotted, rectangular, square] &   1,164 \\ \hline
      size &                                                                                                                            [bulky, long, thin, large, big, tall, short, small, huge, tiny, giant, little, chubby, pudgy] &  16,575 \\ \hline
   texture &                                                                                                                                                             [smooth, fluffy, fuzzy, dry, wet, rusty, bald, hairy, stony] &   1,090 \\ \hline
visibility &                                                                [shiny, unclear, sun, nightime, blurry, shadowy, lit, shady, light, darkened, hazy, dark, barren, cloudy, clear, sunlit, bright, foggy, rainy, sparkling] &  10,454 \\ \hline

\end{tabular}}

    \label{tab:attribute_category_mappings}
\end{table*}

\begin{table*}[t!]
    \centering
    \caption{\textbf{Mean image, text and group scores} for each category of attributes for each model. 
    }
    \resizebox{\textwidth}{!}{\begin{tabular}{l|lll|lll|lll|lll|lll|lll|lll|lll|lll|lll}
\multirow{2}{*}{Model} &                  \multicolumn{3}{c}{age} &   \multicolumn{3}{c}{color} & \multicolumn{3}{c}{expression} & \multicolumn{3}{c}{gender} &  \multicolumn{3}{c}{material}  \\
\cline{2-16}
& Image & Text & Group & Image & Text & Group & Image & Text & Group & Image & Text & Group & Image & Text & Group   \\
\hline
CLIP RN50         &              $47.74$ &              $68.84$ &              $39.20$ &              $18.18$ &              $13.64$ &               $0.00$ &                $21.07$ &               $35.45$ &              $13.14$ &              $34.71$ &              $60.75$ &              $30.18$ &                $0.00$ &                $0.00$ &                $0.00$ \\
CLIP RN50x64      &              $53.77$ &  $\underline{82.41}$ &              $49.25$ &               $4.55$ &              $13.64$ &               $0.00$ &                $20.08$ &               $36.28$ &              $13.97$ &              $39.64$ &     $\mathbf{70.61}$ &              $36.29$ &              $100.00$ &              $100.00$ &              $100.00$ \\
CLIP RN101        &              $48.74$ &              $70.35$ &              $39.20$ &              $18.18$ &              $63.64$ &              $18.18$ &    $\underline{37.60}$ &               $43.97$ &              $24.30$ &              $32.35$ &              $55.23$ &              $27.02$ &                $0.00$ &              $100.00$ &                $0.00$ \\
CLIP ViT          &              $53.27$ &              $71.36$ &              $46.73$ &              $18.18$ &              $77.27$ &              $18.18$ &                $20.91$ &               $38.26$ &              $15.62$ &              $45.17$ &              $69.03$ &              $40.43$ &              $100.00$ &                $0.00$ &                $0.00$ \\
CLIP ViT-B/16     &              $45.73$ &              $76.88$ &              $40.70$ &              $27.27$ &              $77.27$ &              $22.73$ &                $28.84$ &      $\mathbf{66.94}$ &              $24.05$ &              $41.03$ &              $62.92$ &              $35.90$ &                $0.00$ &                $0.00$ &                $0.00$ \\
CLIP ViT L/14     &              $53.77$ &              $75.88$ &              $47.74$ &              $31.82$ &              $72.73$ &              $27.27$ &                $18.43$ &               $16.28$ &              $11.82$ &              $41.62$ &  $\underline{69.43}$ &              $37.67$ &              $100.00$ &                $0.00$ &                $0.00$ \\
CLIP CNN          &              $57.29$ &              $76.38$ &  $\underline{49.75}$ &              $31.82$ &              $68.18$ &              $31.82$ &                $25.21$ &               $47.93$ &              $19.01$ &              $37.67$ &              $56.80$ &              $30.77$ &                $0.00$ &  $\underline{100.00}$ &                $0.00$ \\
CLIP Patched (T)  &              $41.71$ &              $79.40$ &              $37.69$ &              $31.82$ &              $68.18$ &              $31.82$ &                $37.27$ &   $\underline{61.98}$ &     $\mathbf{31.74}$ &              $46.94$ &              $54.83$ &              $36.09$ &              $100.00$ &              $100.00$ &  $\underline{100.00}$ \\
CLIP ViT/B-32     &              $40.20$ &              $74.37$ &              $35.68$ &              $18.18$ &              $13.64$ &               $4.55$ &                $23.22$ &               $58.02$ &              $20.08$ &              $39.05$ &              $60.75$ &              $33.93$ &              $100.00$ &              $100.00$ &              $100.00$ \\
CLIP Patched (V)  &              $44.72$ &              $76.38$ &              $39.70$ &              $40.91$ &  $\underline{90.91}$ &              $40.91$ &                $31.65$ &               $60.74$ &              $27.19$ &              $53.06$ &              $60.16$ &              $43.20$ &              $100.00$ &              $100.00$ &              $100.00$ \\
CLIP Patched (VT) &              $48.74$ &              $72.86$ &              $42.21$ &              $50.00$ &              $81.82$ &              $40.91$ &                $34.55$ &               $61.57$ &  $\underline{29.67}$ &              $52.47$ &              $57.00$ &              $42.60$ &  $\underline{100.00}$ &                $0.00$ &                $0.00$ \\
FLAVA             &     $\mathbf{73.87}$ &              $47.24$ &              $43.22$ &              $86.36$ &              $81.82$ &              $68.18$ &       $\mathbf{39.67}$ &               $24.88$ &              $10.33$ &     $\mathbf{65.09}$ &              $42.60$ &              $37.67$ &              $100.00$ &              $100.00$ &              $100.00$ \\
CLIP RN50x16      &              $52.76$ &     $\mathbf{83.92}$ &              $49.75$ &               $9.09$ &              $13.64$ &               $9.09$ &                $21.98$ &               $42.98$ &              $14.71$ &              $40.83$ &              $62.72$ &              $35.11$ &              $100.00$ &              $100.00$ &              $100.00$ \\
BridgeTower       &  $\underline{68.34}$ &              $80.40$ &     $\mathbf{65.33}$ &     $\mathbf{95.45}$ &              $90.91$ &     $\mathbf{90.91}$ &                $32.23$ &               $52.07$ &              $24.79$ &  $\underline{57.79}$ &              $64.89$ &  $\underline{49.90}$ &              $100.00$ &              $100.00$ &              $100.00$ \\
ViLT              &              $38.19$ &              $53.27$ &              $28.64$ &  $\underline{90.91}$ &     $\mathbf{95.45}$ &  $\underline{86.36}$ &                 $4.79$ &               $18.43$ &               $1.57$ &              $57.59$ &              $67.46$ &     $\mathbf{51.28}$ &     $\mathbf{100.00}$ &     $\mathbf{100.00}$ &     $\mathbf{100.00}$ \\
\hline
\hline
 \multirow{2}{*}{Model} &  \multicolumn{3}{c}{pattern} &  \multicolumn{3}{c}{shape} &   \multicolumn{3}{c}{size} &  \multicolumn{3}{c}{texture} & \multicolumn{3}{c}{visibility} \\
\cline{2-16}
& Image & Text & Group & Image & Text & Group & Image & Text & Group & Image & Text & Group & Image & Text & Group \\ \hline
CLIP RN50         &              $100.00$ &              $100.00$ &              $100.00$ &               $3.69$ &              $23.88$ &              $2.56$ &              $21.21$ &              $47.95$ &              $17.11$ &              $0.00$ &              $20.69$ &              $0.00$ &                 $0.00$ &                $0.00$ &              $0.00$ \\
CLIP RN50x64      &              $100.00$ &              $100.00$ &              $100.00$ &               $9.46$ &              $22.92$ &              $6.25$ &              $18.83$ &              $33.40$ &              $12.37$ &              $3.45$ &              $20.69$ &              $0.00$ &                 $0.00$ &                $0.00$ &              $0.00$ \\
CLIP RN101        &              $100.00$ &              $100.00$ &              $100.00$ &               $9.13$ &              $27.88$ &              $5.45$ &              $20.96$ &              $37.63$ &              $12.72$ &              $3.45$ &     $\mathbf{31.03}$ &              $3.45$ &                $50.00$ &                $0.00$ &              $0.00$ \\
CLIP ViT          &              $100.00$ &              $100.00$ &              $100.00$ &              $10.26$ &  $\underline{38.94}$ &              $6.41$ &              $24.13$ &              $42.98$ &              $18.85$ &  $\underline{6.90}$ &               $6.90$ &              $0.00$ &                $50.00$ &                $0.00$ &              $0.00$ \\
CLIP ViT-B/16     &                $0.00$ &              $100.00$ &                $0.00$ &               $8.01$ &              $21.63$ &              $3.21$ &              $15.69$ &              $28.33$ &               $7.73$ &              $6.90$ &              $17.24$ &              $0.00$ &      $\mathbf{100.00}$ &                $0.00$ &              $0.00$ \\
CLIP ViT L/14     &                $0.00$ &              $100.00$ &                $0.00$ &  $\underline{13.46}$ &              $37.02$ &  $\underline{8.49}$ &              $24.84$ &              $45.11$ &  $\underline{19.34}$ &              $3.45$ &               $6.90$ &              $0.00$ &                $50.00$ &                $0.00$ &              $0.00$ \\
CLIP CNN          &  $\underline{100.00}$ &  $\underline{100.00}$ &  $\underline{100.00}$ &              $12.82$ &     $\mathbf{39.58}$ &     $\mathbf{8.81}$ &              $17.51$ &              $33.43$ &              $10.62$ &              $0.00$ &               $6.90$ &              $0.00$ &                 $0.00$ &                $0.00$ &              $0.00$ \\
CLIP Patched (T)  &                $0.00$ &                $0.00$ &                $0.00$ &               $6.25$ &              $22.92$ &              $3.04$ &              $20.40$ &              $43.08$ &              $16.09$ &              $3.45$ &              $10.34$ &              $0.00$ &                $50.00$ &               $50.00$ &              $0.00$ \\
CLIP ViT/B-32     &                $0.00$ &                $0.00$ &                $0.00$ &               $7.85$ &              $26.60$ &              $4.97$ &              $13.63$ &              $34.47$ &               $7.05$ &              $0.00$ &               $3.45$ &              $0.00$ &                $50.00$ &                $0.00$ &              $0.00$ \\
CLIP Patched (V)  &                $0.00$ &              $100.00$ &                $0.00$ &              $10.10$ &              $28.04$ &              $6.41$ &              $16.29$ &              $44.60$ &              $10.42$ &              $0.00$ &               $6.90$ &              $0.00$ &    $\underline{50.00}$ &   $\underline{50.00}$ &    $\mathbf{50.00}$ \\
CLIP Patched (VT) &                $0.00$ &                $0.00$ &                $0.00$ &               $7.37$ &              $23.24$ &              $5.45$ &              $25.14$ &              $45.67$ &              $17.03$ &              $3.45$ &  $\underline{27.59}$ &              $3.45$ &                 $0.00$ &               $50.00$ &              $0.00$ \\
FLAVA             &                $0.00$ &                $0.00$ &                $0.00$ &     $\mathbf{34.62}$ &               $3.37$ &              $2.72$ &     $\mathbf{48.66}$ &              $16.09$ &              $12.60$ &    $\mathbf{31.03}$ &               $3.45$ &  $\underline{3.45}$ &                 $0.00$ &                $0.00$ &              $0.00$ \\
CLIP RN50x16      &     $\mathbf{100.00}$ &              $100.00$ &     $\mathbf{100.00}$ &               $7.85$ &              $19.87$ &              $4.49$ &              $22.93$ &     $\mathbf{54.13}$ &              $18.88$ &              $3.45$ &              $20.69$ &              $0.00$ &                 $0.00$ &                $0.00$ &              $0.00$ \\
BridgeTower       &              $100.00$ &              $100.00$ &              $100.00$ &               $7.53$ &              $32.53$ &              $6.41$ &  $\underline{25.54}$ &  $\underline{49.06}$ &     $\mathbf{20.40}$ &              $0.00$ &              $17.24$ &              $0.00$ &                 $0.00$ &     $\mathbf{100.00}$ &              $0.00$ \\
ViLT              &                $0.00$ &     $\mathbf{100.00}$ &                $0.00$ &               $5.13$ &              $20.99$ &              $2.56$ &              $17.44$ &              $45.36$ &              $14.75$ &              $6.90$ &              $27.59$ &     $\mathbf{3.45}$ &                 $0.00$ &                $0.00$ &  $\underline{0.00}$ \\
\end{tabular}}

    \label{tab:my_label}
\end{table*}

\begin{figure*}[t!]
    \centering
    \includegraphics[width=\linewidth]{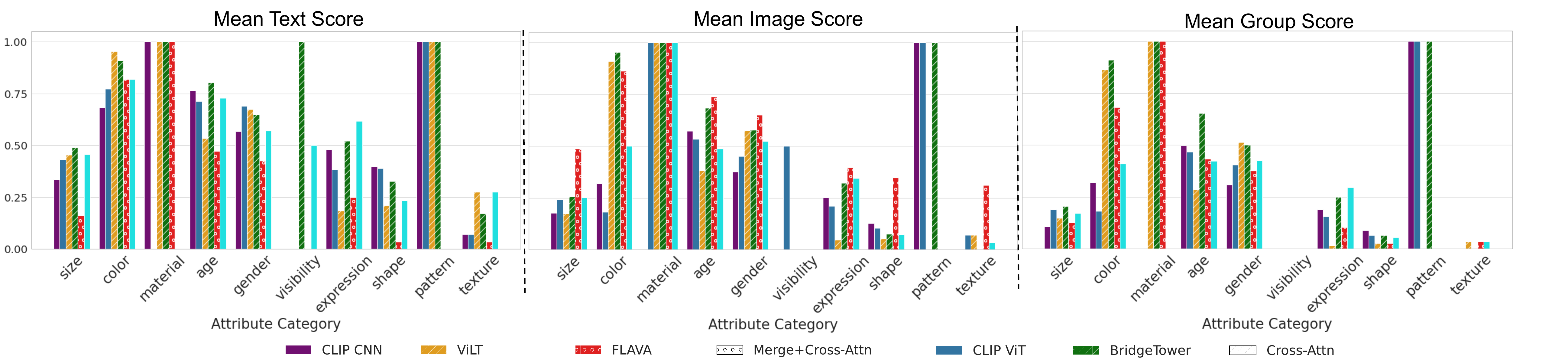}

    \caption{\textbf{Performance on compositional understanding}Mean image, text and group scores for a subset of models. Models are typically better matching a caption given an image rather than the reverse.}
    \label{fig:attribute_barplot}
\end{figure*}

\begin{table*}[t!]
    \centering
    \caption{\textbf{The objects that belong to each category for the object-context analysis} on specific objects in Probe-B.}
    \resizebox{\textwidth}{!}{\begin{tabular}{l|l|l}
Object & Category & Groups\\
\hline
          accessories &                                                                      [backpack, umbrella, handbag, tie, suitcase] &    174 \\ \hline
         animals &                                               [bird, cat, dog, horse, sheep, cow, elephant, bear, zebra, giraffe] &    627 \\ \hline
      appliances &                                                                          [microwave, oven, toaster, refrigerator] &    591 \\ \hline
           decor &                                                                                                     [clock, vase] &    138 \\ \hline
     electronics &                                                                 [tv, laptop, mouse, remote, keyboard, cell phone] &   1095 \\ \hline
        fixtures &                                                                                                    [toilet, sink] &    387 \\ \hline
           foods &                                                                           [sandwich, hot dog, pizza, donut, cake] &    258 \\ \hline
          fruits &                                                                                                  [banana, orange] &    120 \\ \hline
       furniture &                                                                                 [chair, couch, bed, dining table] &    546 \\ \hline
     kitchenware &                                                               [bottle, wine glass, cup, fork, knife, spoon, bowl] &    399 \\ \hline
          people &                                                                                                          [person] &    720 \\ \hline
          plants &                                                                                                    [potted plant] &    108 \\ \hline
      recreation & [frisbee, skis, snowboard, sports ball, kite, baseball bat, baseball glove, skateboard, surfboard, tennis racket] &    117 \\ \hline
         roadway &                                                           [traffic light, fire hydrant, stop sign, parking meter] &    144 \\ \hline
street furniture &                                                                                                           [bench] &     42 \\ \hline
           tools &                                                                                [scissors, hair drier, toothbrush] &     15 \\ \hline
            toys &                                                                                                [book, teddy bear] &    144 \\ \hline
      vegetables &                                                                                                [broccoli, carrot] &    111 \\ \hline
        vehicles &                                                     [bicycle, car, motorcycle, airplane, bus, train, truck, boat] &    603 \\ \hline

\end{tabular}}

    \label{tab:object_category_mappings}
\end{table*}

\begin{figure*}[t!]
    \centering
    \includegraphics[width=0.95\linewidth]{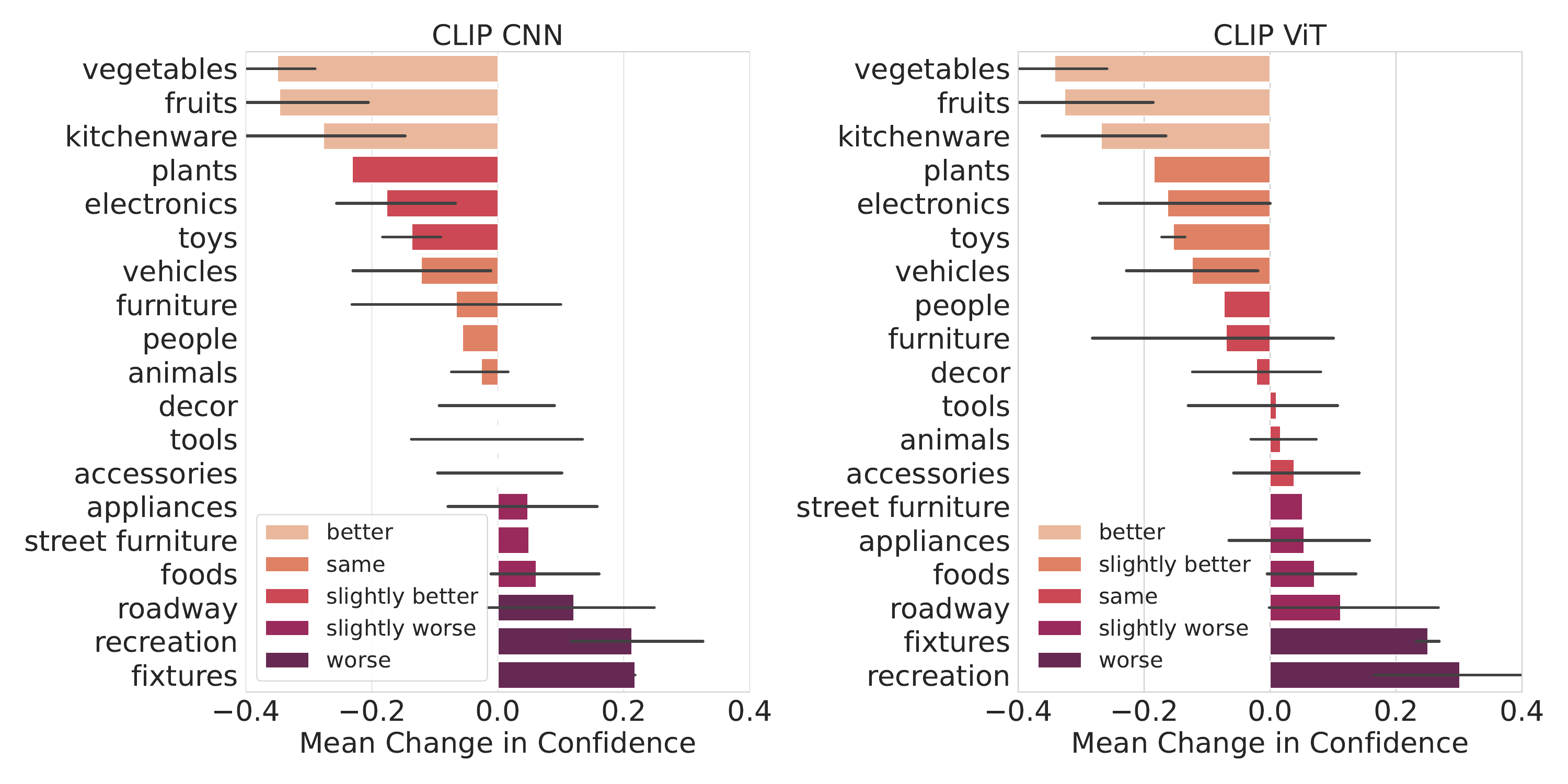}
    \caption{\textbf{Comparing the change in confidence} from a patched image $\tilde{x_0}$ to the image where all other objects and background $\tilde{x_1}$ is removed aggregated over CLIP backbones.}
    \label{fig:cooccurence_cnn_vit}
\end{figure*}

\begin{table*}[t!]
    \centering
    \caption{\textbf{Mean results for Probe-$B_{MR}$} when the background of an image is replaced with each filler (top) and for each model averaged over fillers (bottom). Comparisons are between the original image $x_0$, original image with a random patch $\tilde{x_0}$ and the modified image $\tilde{x_1}$ where the background is removed. The metrics are mean average precision (mAP), relative robustness ($\gamma_r$) measuring the relative drop/increase in performance, and mean change in softmax confidence $\mu(\triangledown(c))$ for the objects.
}
    \resizebox{\linewidth}{!}{\begin{tabular}{|l|r|r|r|r|r|r|r|r|r|}

 & \multicolumn{3}{c|}{Average Precision (mAP)} & \multicolumn{3}{c|}{Relative Robustness ($\gamma_r$)} &  \multicolumn{3}{c|}{ Mean Change Confidence ($\mu(\triangledown (c)$)} \\
 \cline{2-10}
Filler &  $x_0$ &  $\tilde{x_0}$ &  $\tilde{x_1}$ & $(x_0, \tilde{x_1})$ & $(x_0, \tilde{x_0})$ &  $(\tilde{x_0}, \tilde{x_1})$ &    $(c_0 - \tilde{c_0})$ & $(\tilde{c_0} - \tilde{c_1})$ & $(c_0 - \tilde{c_1})$ \\
\hline
black & $69.76$ & $\underline{69.62}$ & $\underline{70.6}$  & $\underline{1.09}$ & $\mathbf{1.05}$ & $\underline{1.18}$ & $\underline{0.95}$ & $\underline{-0.91}$ & $\mathbf{0.04}$ \\
noise & $\underline{69.8}$  & $67.75$ & $68.14$ & $1.04$ & $\underline{1.02}$ & $1.2$  & $1.97$ & $\underline{-0.91}$ & $1.06$ \\
gray  & $\mathbf{69.84}$ & $\mathbf{69.68}$ & $\mathbf{71.02}$ & $\mathbf{1.1}$  & $\mathbf{1.05}$ & $1.2$  & $\mathbf{0.92}$ & $\mathbf{-0.86}$ & $\underline{0.05}$ \\
scene & $\underline{69.8}$  & $66.5$  & $67.85$ & $1.04$ & $1$    & $\mathbf{1.29}$ & $2.18$ & $-1.31$ & $0.87$ \\
\hline
\hline
Model &  $x_0$ &  $\tilde{x_0}$ &  $\tilde{x_1}$ &  $(x_0, \tilde{x_1})$ & $(x_0, \tilde{x_0})$ &  $(\tilde{x_0}, \tilde{x_1})$ &   $(c_0 - \tilde{c_0})$ & $(\tilde{c_0} - \tilde{c_1})$ & $(c_0 - \tilde{c_1})$ \\
\hline
CLIP RN50     &              $65.05$ &                 $65.47$ &                 $60.30$ &                       $1.00$ &                       $1.02$ &                               $0.99$ &                      $-0.05$ &                                           $4.86$ &                                   $4.82$ \\
CLIP ViT/B-32 &              $68.77$ &                 $67.49$ &                 $61.10$ &                       $0.95$ &                       $0.99$ &                               $0.98$ &                                   $0.40$ &                                           $5.12$ &                                   $5.52$ \\
CLIP CNN      &              $63.23$ &                 $63.56$ &                 $63.46$ &              $\underline{1.12}$ &                       $1.02$ &                               $1.10$ &                                   $0.46$ &                                           $2.64$ &                                   $3.11$ \\
CLIP RN101    &              $64.56$ &                 $65.00$ &                 $63.80$ &           $1.09$ &                       $1.02$ &                               $1.08$ &                                   $0.21$ &                                           $2.79$ &                                   $3.00$ \\
CLIP ViT-B/16 &              $69.97$ &                 $68.54$ &                 $65.15$ &                       $0.99$ &                       $0.98$ &                               $1.02$ &                                   $0.36$ &                                           $2.92$ &                                   $3.28$ \\
FLAVA         &              $72.05$ &     $74.47$ &                 $66.75$ &                       $0.98$ &           $1.05$ &                               $0.94$ &                                   $0.00$ &                                           $0.01$ &                       $0.01$ \\
CLIP ViT L/14 &              $70.98$ &                 $69.38$ &                 $68.99$ &                       $1.04$ &                       $0.98$ &                               $1.08$ &                                   $0.94$ &                                           $1.17$ &                                   $2.12$ \\
CLIP ViT      &              $71.05$ &                 $70.94$ &                 $71.50$ &                       $1.08$ &                       $1.01$ &                               $1.08$ &                                   $0.70$ &                                           $0.52$ &                                   $1.22$ \\
ViLT          &     $\mathbf{83.49}$ &                 $71.38$ &     $\underline{83.26}$ &                       $1.00$ &                       $0.87$ &                      $\mathbf{1.62}$ &                                   $6.68$ &                                 $\mathbf{-6.61}$ &                                   $0.08$ \\
BridgeTower   &  $\underline{81.85}$ &        $\mathbf{81.88}$ &        $\mathbf{83.40}$ &                       $1.05$ &              $1.06$ &                   $1.23$ &                         $\mathbf{-0.79}$ &                              $-0.42$ &                         $\mathbf{-1.22}$ \\
BLIP     & $68.67$ & $73.43$ & $71.76$ & $1.07$ & $\mathbf{1.16}$ & $1.08$ & $\underline{-0.78}$ & $0.16$  & $\underline{-0.62}$ \\
BLIP2    & $77.37$ & $\underline{76.19}$ & $77.1 $ & $1.03$ & $1.05$ & $1.17$ & $3.47$  & $-1.96$ & $1.51$  \\
OTTER    & $30.75$ & $29.23$ & $30.8 $ & $1.05$ & $1.08$ & $1.3 $ & $0  $   & $0  $   & $0 $    \\
ALIGN    & $50.53$ & $50.52$ & $48.89$ & $\mathbf{1.48}$ & $\mathbf{1.15}$ & $\underline{1.46}$ & $0  $   & $0.04$  & $0.04$  \\
MetaCLIP & $67.1 $ & $61.79$ & $66.76$ & $1   $ & $0.99$ & $1.43$ & $5.22$  & $\underline{-4.86}$ & $0.36$  \\
SigLIP   & $69.96$ & $70.46$ & $69.77$ & $1   $ & $1.08$ & $1.16$ & $2.19$  & $-2.07$ & $0.13$  \\
\end{tabular}}

\label{tab:context}
\end{table*}

\begin{table*}[t!]
\centering
\caption{\textbf{Results for when the background and all other objects are replaced} with a filler $\tilde{x_1}$, compared to the original image $x_0$, and an image with a random patch of the same filler type $\tilde{x_0}$. Metrics used are the accuracy of detecting the object compared to other objects that are not present in the image and the relative robustness $\gamma_r$, which is the relative change in confidence.
}
\resizebox{.75\textwidth}{!}{\begin{tabular}{l|lll|lll|}
& \multicolumn{3}{c|}{Accuracy} & \multicolumn{3}{c|}{Relative Robustness $\gamma^r$} \\
\cline{2-7}
 Filler &$x_0$ & $\tilde{x_0}$ & $\tilde{x_1}$ & $(x_0, \tilde{x_1})$ & $(x_0, \tilde{x_0})$ & $(\tilde{x_0}, \tilde{x_1})$ \\

\hline
noise & $54.48$ & $54.95$ & $60.09$ & $1.1 $ & $\underline{1.01}$ & $1.09$ \\
scene & $49.08$ & $52   $ & $56.1 $ & $\mathbf{1.14}$ & $\mathbf{1.06}$ & $1.08$ \\
black & $\underline{56.62}$ & $\mathbf{57.13}$ & $\underline{63.79}$ & $\underline{1.13}$ & $\underline{1.01}$ & $\underline{1.12}$ \\
gray  & $\mathbf{57.08}$ & $\underline{56.83}$ & $\mathbf{63.98}$ & $1.12$ & $1   $ & $\mathbf{1.13}$ \\
\hline
\hline
 Model &$x_0$ & $\tilde{x_0}$ & $\tilde{x_1}$ & $(x_0, \tilde{x_1})$ & $(x_0, \tilde{x_0})$ & $(\tilde{x_0}, \tilde{x_1})$ \\
 \hline
BridgeTower   &     $\mathbf{77.36}$ &        $\mathbf{76.14}$ &      $\mathbf{77.70}$ &                            $1.01$ &                              $0.98$ &                               $1.02$ \\
FLAVA         &              $56.33$ &     $59.16$ &               $58.32$ &                            $1.03$ &                              $1.06$ &                               $0.99$ \\
CLIP ViT/B-32 &              $46.19$ &                 $50.50$ &               $54.20$ &                            $1.17$ &                     $1.10$ &                               $1.07$ \\
CLIP ViT-B/16 &              $51.12$ &                 $51.31$ &               $60.30$ &                            $1.18$ &                              $1.01$ &                               $1.17$ \\
CLIP ViT L/14 &              $56.27$ &                 $54.18$ &               $67.16$ &                            $1.19$ &                              $0.96$ &                               $1.24$ \\
CLIP RN50     &              $45.79$ &                 $48.64$ &               $55.68$ &                            $1.21$ &                  $1.07$ &                               $1.14$ \\
CLIP ViT      &  $59.31$ &                 $56.08$ &               $71.76$ &                            $1.21$ &                              $0.95$ &                               $\underline{1.28}$ \\
CLIP RN101    &              $46.48$ &                 $47.92$ &               $57.32$ &                            $1.23$ &                              $1.03$ &                               $1.20$ \\
CLIP RN50x16  &              $53.88$ &                 $50.92$ &               $66.18$ &                            $1.23$ &                              $0.95$ &                   $\mathbf{1.30}$ \\
CLIP RN50x64  &              $56.86$ &                 $53.70$ &               $70.04$ &                            $1.23$ &                              $0.95$ &                               $\mathbf{1.30}$ \\
CLIP CNN      &              $50.08$ &                 $49.66$ &               $62.18$ &                $\underline{1.24}$ &                              $1.00$ &                               $1.25$ \\
ViLT          &              $54.73$ &                 $55.82$ &   $\underline{72.09}$ &                   $\mathbf{1.33}$ &                              $1.02$ &                      $\mathbf{1.30}$ \\
BLIP     & $\underline{68.41}$ & $\underline{73.95}$ & $64.62$ & $0.94$ & $1.084$ & $0.87$ \\
BLIP2    & $68.02$ & $65.13$ & $65.67$ & $0.97$ & $0.964$ & $1.01$ \\
OTTER    & $12.16$ & $10.45$ & $11.89$ & $0.98$ & $0.864$ & $1.14$ \\
ALIGN    & $63.12$ & $61.66$ & $69.17$ & $1.1 $ & $0.984$ & $1.12$ \\
MetaCLIP & $48.01$ & $55.03$ & $51.37$ & $1.07$ & $\underline{1.154}$ & $0.93$ \\
SigLIP   & $63.51$ & $73.82$ & $62.22$ & $0.98$ & $\mathbf{1.164}$ & $0.84$ \\
\end{tabular}} 

\label{tab:cooccurence_agg}
\end{table*}

\begin{table*}[t!]
\centering
\caption{\textbf{The results for varying the alpha values} for patching \cite{ilharco2022patching} finetuned CLIP models on either text encoder (T), visual encoder (V), or both (VT). There is a clear trade-off with downstream ImageNet classification and finetuning on a smaller, compositional and relational focused dataset.}
    \resizebox{0.85\textwidth}{!}{
\begin{tabular}{l|l|lll|ll}
  &   &  \multicolumn{3}{c|}{RelComp}&  \multicolumn{2}{c}{ImageNet} \\
 \cline{1-7}
 Stream & alpha & Group Score &   Image Score  &  Text Score &  Top1              & Top5 \\
 
\hline
     v &    0.2 &             $31.52$ &             $54.67$ &             $53.17$ &    $\mathbf{61.45}$ &    $\mathbf{87.73}$ \\
     v &    0.3 &             $32.58$ &             $55.61$ &             $53.97$ & $\underline{58.25}$ & $\underline{85.69}$ \\
     v &    0.4 &             $33.29$ &             $56.40$ &             $54.89$ &             $54.19$ &             $82.72$ \\
     v &    0.5 &             $34.15$ &             $57.31$ &             $55.60$ &             $49.36$ &             $78.87$ \\
     v &    0.6 &             $34.43$ &             $57.62$ &             $55.94$ &             $44.04$ &             $73.96$ \\
    vt &    0.2 &             $42.19$ &             $64.30$ &             $62.45$ &             $54.62$ &             $83.05$ \\
     t &    0.2 &             $47.18$ &             $67.86$ &             $66.62$ &             $57.42$ &             $85.14$ \\
    vt &    0.3 &             $49.58$ &             $69.83$ &             $68.00$ &             $44.92$ &             $73.93$ \\
    vt &    0.4 &             $50.11$ &             $70.14$ &             $68.78$ &             $34.95$ &             $63.07$ \\
    vt &    0.5 &             $51.57$ &             $71.12$ &             $70.35$ &             $27.00$ &             $52.54$ \\
    vt &    0.6 &             $53.38$ &             $72.62$ &             $71.77$ &             $20.98$ &             $43.88$ \\
     t &    0.3 &             $56.31$ &             $74.56$ &             $73.55$ &             $50.69$ &             $79.35$ \\
     t &    0.4 &             $62.51$ &             $78.57$ &             $78.29$ &             $44.39$ &             $72.39$ \\
     t &    0.5 & $\underline{70.36}$ & $\underline{83.55}$ & $\underline{83.21}$ &             $38.66$ &             $66.23$ \\
     t &    0.6 &    $\mathbf{74.63}$ &    $\mathbf{86.26}$ &    $\mathbf{85.62}$ &             $33.54$ &             $60.21$ \\
\hline
\end{tabular}}

\label{tab:patch_ablation}
\end{table*}

\begin{table*}[t!]
    \centering
    \caption{\textbf{The pre-training datasets} include MSCOCO \cite{coco}, SBU Captions, Localized Narratives (LN), Visual Genome (VG) \cite{visualgenome}, Wikipedia Image Text (WIT) \cite{srinivasan2021wit}, Conceptual Captions (CC) \cite{sharma-etal-2018-conceptual}, Conceptual Captions 12M (CC12) \cite{changpinyo2021conceptual}, Red Caps (RC) \cite{desai2021redcaps}, YFCC100M \cite{yfcc}, and LAION-400M \cite{schuhmann2021laion}.
    }
    \resizebox{\textwidth}{!}{\begin{tabular}{l|c|c|c|c|c|p{3cm}|}
    Model & Params &  Datasets & Images  & Captions & Arch. & Attn \\
    \hline
      CLIP RN50 \cite{clip} & 102M  & LAION-400M  &  400M & 400M & dual-stream & modality-specific \\ \hline
      CLIP RN101 \cite{clip} & 121M & LAION-400M  &  400M & 400M & dual-stream & modality-specific \\ \hline
      CLIP ViT B16/32 \cite{clip} & ~150M & LAION-400M  &  400M & 400M & dual-stream & modality-specific \\ \hline
      CLIP ViT L14 \cite{clip} & 428M & LAION-400M  &  400M & 400M & dual-stream & modality-specific \\ \hline
      FLAVA \cite{singh2022flava} & 358M & MSCOCO, SBU, LN, CC, CC12, VG, WIT, RC, YFCC100M  & 70M & 70M & dual-stream & modality-specific, merged \\ \hline
      ViLT \cite{kim2021vilt} & 112M & MSCOCO,VG,SBU,CC & 4.20M & 9.58M & single-stream & modality-specific, merged \\ \hline
      Bridgetower \cite{xu2022bridge}& 865M & MSCOCO,VG,SBU,CC & 4.20M & 9.58M & dual-stream & modality-specific, co-attn, merged\\ 
    \end{tabular}}
    \label{tab:models_and_data}
\end{table*}

\begin{figure*}[t!]
\centering
    \includegraphics[width=\linewidth]{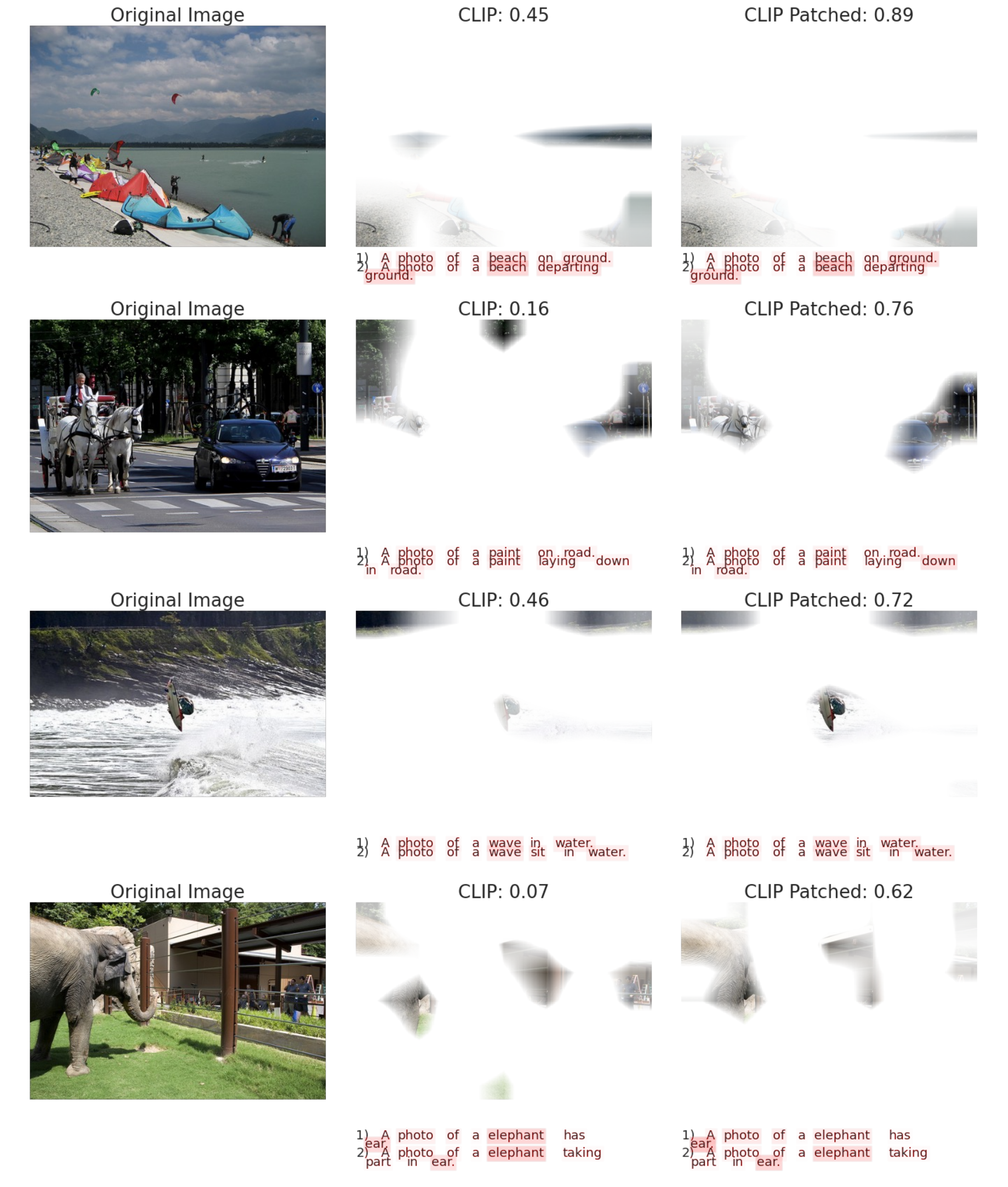}
    \caption{\textbf{Examples from Probe-R} comparing CLIP ViT-B/32 to the same model finetuned on RelComp for both the visual and text encoder, then patched \cite{ilharco2022patching}. The values are the softmax confidence for the correct prompt $P_{R_1}$ shown as $1)$ vs the incorrect prompt $2)$, where the predicate is swapped, or $P_{R_2}$.}
    \label{fig:clip_patched}
\end{figure*}

\begin{figure*}[t!]
\centering
    \includegraphics[width=\linewidth]{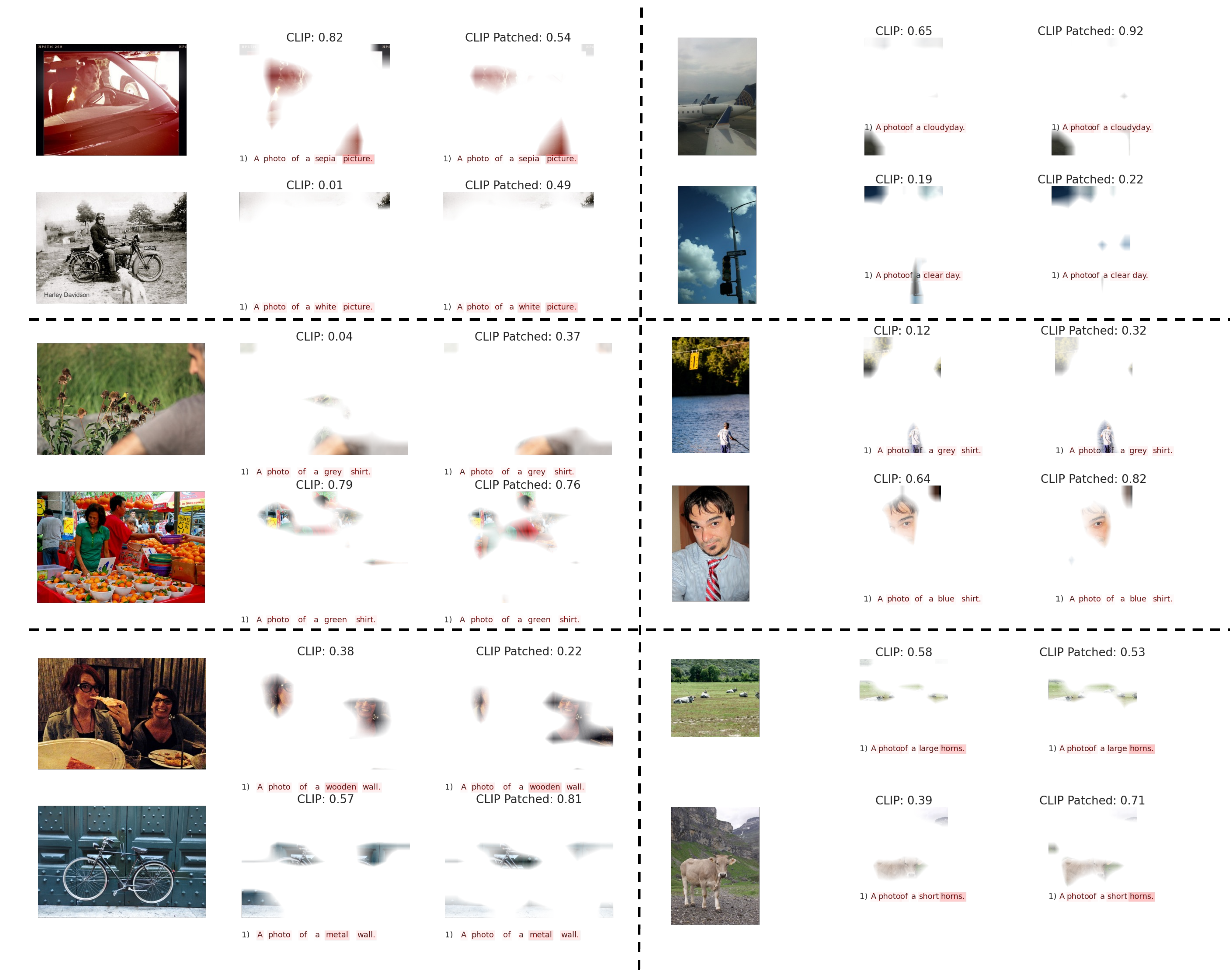}
    \caption{\textbf{Examples from Probe-C} comparing CLIP ViT-B/32 to the same model finetuned on RelComp for both the visual and text encoder, then patched \cite{ilharco2022patching}. For each group, the first image and its corresponding prompt are on top, and the second image and prompt are on the bottom. The values are the softmax confidence for the corresponding prompt when compared to the alternative prompt. }
    \label{fig:clip_patched_comp}
\end{figure*}

\begin{figure}[t!]
   \centering
    \includegraphics[width=\linewidth]{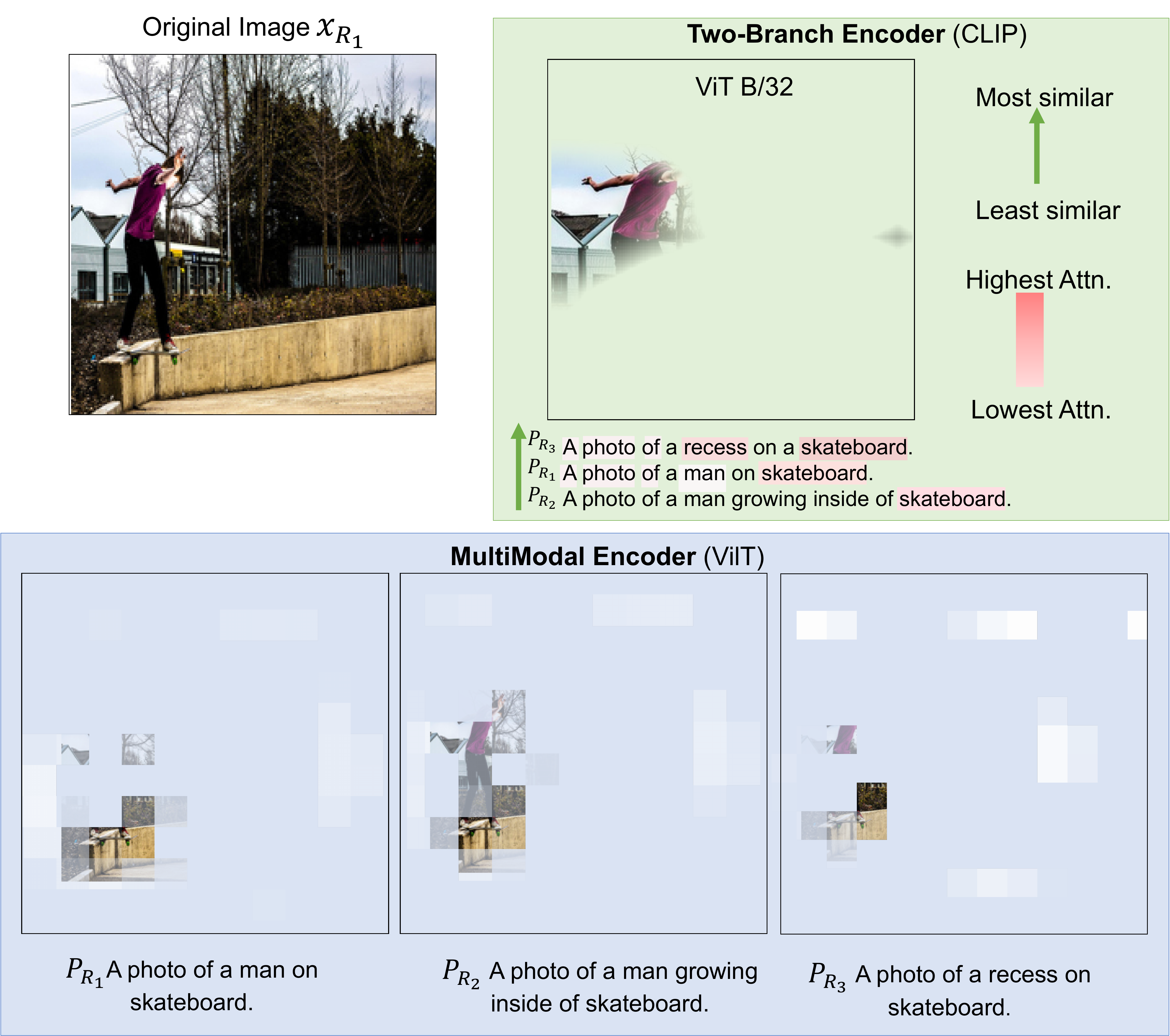}
    \caption{We design probes that measure \textbf{relational understanding} in V+L models, in this case we compare ViLT \cite{kim2021vilt} that uses cross-attention (top) and CLIP \cite{clip} which does not (bottom). With cross-attention, the model can change its focus based on the prompt and performs better when compositions and relations are swapped for unrealistic/non-present ones. Meanwhile, CLIP does not adapt and focuses more highly on objects, like ``man'' and ``skateboard''. 
    }
    \label{fig:clip_teaser}

\end{figure}

The results for all models for the Probe-R benchmark are shown in Table \ref{tab:sup_relation_eval}. We include CLIP models we finetuned on RelComp, training either the text encoder (T), visual encoder (V) or both encoders (VT). Training only the text encoder seems to have the highest improvement, but as mentioned in the paper, the largest occurrence of ``catastrophic forgetting'' when evaluated on ImageNet. A TSNE plot of model features that includes CLIP Patched (VT) is shown in Figure \ref{fig:relation_tsne}. In black we have the image features, in red we have the predicate swapped text features ($P_{R_2}$), and in green we have the ground truth relation text features ($P_{R_1}$). This finetuned and patched version appears to have tighter clusters compared to the original CLIP model. 

\subsection{Probe-C: Compositional Understanding}
This dataset was generated using MSCOCO \cite{coco}. To guarantee that the images had no similarity or overlap, we focused on using antonyms of select attributes. We started by using NLTK POS \cite{bird2009natural} to find adjective-noun pairs. We then manually cleaned and extracted the adjectives to guarantee the attribute is a visual one such as ``red'' or ``young'' as opposed to a subjective one such as ``hungry'' or ``thirsty''. While these are useful attributes, we are primarily interested in visual perception as opposed to subjective inference. We then iterated through all images and mapped each attribute to their corresponding image IDs, and we did the same with objects. Using this collection, we were able to create groups of pairs based on either swapping the attribute to one of its antonyms or swapping the object with one that has the same attribute. 

The overall results for Probe-C for all models is in Table \ref{tab:compositional_eval}. The mappings we used to categorize different attributes is shown in Table \ref{tab:attribute_category_mappings}, these were manually generated. A visual break down of different model performances for each attribute is shown in Figure \ref{fig:attribute_barplot}. From there, you can see the changes in score based on whether it is matching the caption given the image versus given text. We also see that most models struggle with ``visibility'' and often ``texture''.

\subsection{Probe-B: Context Understanding}
In set 1, for each image we remove the background using segmentation masks from original annotations. We replace the background with 1 of four fillers: black, gray, Gaussian noise, or a random scene. Random scenery was collected from the Indoor Scenes Dataset \cite{quattoni2009recognizing} and the Kaggle Landscape dataset \cite{arnaud58}. These images were manually filtered to ensure none of the 80 MSCOCO classes were present. The total collection is 31,745 images with 4 fillings each for a total of 126,980 images. We filtered images based on a threshold for how much background can be removed to ensure that some context was actually removed. In set 2, for each image we remove all other objects and the background using segmentation masks. In this case, $x_0$ is the image with all objects with just the background removed while $\tilde{x_1}$ is the image with just one object remaining and all other objects and the background removed. This allows us to isolate whether it is the other objects compared to background removal. Like in set 1, we replace them with the different possible fillers. Images are chosen if they do not have overlapping bounding boxes and if their object area is over a threshold to allow for better visibility. Prompts for set 2 only include objects not present in the original image and the target object.

To better compare CLIP backbones, Figure \ref{fig:cooccurence_cnn_vit} shows a comparison between the change in confidence from a patched image $\tilde{x_0}$ to the image where all other objects and background $\tilde{x_1}$ is removed aggregated over CLIP backbones. Table \ref{tab:object_category_mappings} shows what objects are assigned to which category and how many samples are present in the annotations. The main differences are in objects they struggle with by how much and in which order. 

Overall results for Probe-B are in Table \ref{tab:context} and \ref{tab:cooccurence_agg}. In both cases, replacing with scene and noise produces worse results compared to black and gray fillers.  For aggregating across filler, we only include CLIP ViT-L/14@336px, CLIP RN50x4, FLAVA, ViLT, BridgeTower, BLIP, BLIP2, OTTER, ALIGN, MetaCLIP and SigLIP. When comparing individual model results in Table \ref{tab:cooccurence_agg}, performance tends to increase when only the other object remains, meaning that other objects may actually distract models. BridgeTower is the highest performer and has the lowest robustness from $x_0$ to $x_1$ meaning that it may be using some level of object relationship understandings to help recognize objects. However, this difference is minor and therefore inconclusive. Other models' robustness though is higher indicating they perform better when objects are in isolation, indicating they are not using object relationship understanding to help object detection of particular objects. In Table \ref{tab:context}, when only background is removed, we see little change. However, in ViLT, which is one transformer that takes both text and visual tokens, adding a patch reduced performance noticeably worse when compared to other models. This may indicate a weakness in a single-stream, transformer based approach.

\section{Exploring Improving Dual-Stream Only  Conceptual Models}
\label{supp:clip_patched}
Based on our evaluation of these models, we see that cross-attention between modalities improves the learning of conceptual models about objects and actions in a system and the relationships between them. However, a limitation of this approach is its use for downstream tasks. Both ViLT and BridgeTower require image-text pairs of input, making other tasks like image classification \textit{computationally expensive and difficult}. Meanwhile, dual-stream encoders like CLIP and FLAVA allow  uni-modal feature representations that can be extracted and used for a variety of downstream tasks. Improving models that do not require paired input would provide greater value and stronger representations. To explore this idea, we fine-tune CLIP on a new dataset inspired by this benchmark called RelComp. 

\subsection{Method}
In order to improve CLIP for compositional and relational understanding, we propose using selective negative and positive pairing based on compositional and predicate swaps. We propose using two losses, an image-text matching (ITM) loss and a contrastive loss (C) similar to CLIP \cite{clip} and FLAVA \cite{singh2022flava}. The ITM loss is a triplet loss with two instances \cite{chechik2010large}, maximizing the distance between an anchor and a negative sample while minimizing the distance between an anchor and a positive sample. 
We use this in order to focus model learning on compositions and relations. 
 
The first is where the anchor is the image $x$, the positive is the caption $p$, and the negative $\overline{p}$ is the same caption but with either the predicate or the composition swapped. The second uses a real-world caption $y$ as an anchor and the corresponding image $x$ as a positive. The final ITM loss is the average of the two. For the contrastive loss, we maximize the cosine similarities between image and text pairs and minimize those for the image and negative text pairs. We use two versions, the first uses the real-world captions $y$ and their corresponding images, and the second uses the positive text prompts $p$ and their images. The final contrastive loss is the average of the two.  A summary of this approach is shown in Figure \ref{fig:contrastive_training}.

\subsection{Dataset: RelComp}
We used our existing knowledge of the benchmark to generate a new training and testing dataset. For compositions, we use images and captions from the MSCOCO dataset \cite{coco}. For anchor text we use the real-world caption, for positive we replace all compositions with synonyms, and for negatives we replace all compositions with antonyms. No captions seen in this dataset are also seen in Probe-C. For relations, we use images, region descriptions and relationships from the VisualGenome dataset \cite{visualgenome}. For each image, we find the region description that has the most overlap with prompts generated in the same way as Probe-R and use this as our anchor caption. 
For negative, we use the same template but use prompt with the predicate swapped to an unlikely one, as in Probe-R. To prevent exact prompts from the benchmark being included, we filtered for images that are not present in Probe-R. This results in 149,166 groups with 78,155 of those swapping compositions and 71,011 swapping predicates for training. The test set has 15,836 groups and of those, 8,734 are swap compositions and 7,102 swap predicates.

\subsection{Implementation}
We finetune the CLIP ViT-B/32 model using our proposed ITM and contrastive loss on the proposed dataset RelComp. We use stochastic gradient descent with a cosine learning rate scheduler with a minumum learning rate of .001, momentum 0.9, weight decay of .0001. We train for 40 epochs using an 11GB GPU and a batch size of 128.   We use these smaller configurations to show the benefits with just light tuning. One of the many challenges of fine-tuning a large model, is that the distribution shift may lead to a loss of the original feature space. In order to prevent this ``catastrophic forgetting'' of the original feature space, we linearly interpolate the original CLIP weights with our finetuned weights using an alpha$=0.2$, leaning more towards the original weights, in order to reduce this shift \cite{ilharco2022patching,wortsman2022robust}. This is referred to as patching and therefore we call the finetuned and patched version ``CLIP Patched''. We finetune three configurations based on which encoders we finetune: visual only (V), text only (T) or both (VT).

\subsection{Results}
Overall results for our experiment are shown in Table \ref{tab:training_results2}. When finetuning on the new dataset, there is an issue of drift from the original CLIP performance as measured by ImageNet accuracy, even when patching. When finetuning using the visual-encoder only, the drift is less pronounced, but so is the improvement on RelComp. The largest increase in RelComp is seen when just training the text encoder. (1) This may indicate that for non-cross-attention models text is more important for conceptual mapping. Overall, (2) our findings indicate that it is possible by using selective negative sampling to enforce compositional and relational learning without extensive co-attention and computational complexity. Limitations of this experiment is our training data is very small in comparison to recent works, further work should investigate this relationship with a larger dataset with more variation. 
\begin{table}
    \centering
    \caption{\textbf{Overall results for finetuning and patching the CLIP ViT-B/32 on the proposed RelComp dataset}. ImageNet accuracy is shown to measure the drift from the original CLIP space. RelComp is the image score for the correct image-to-prompt matching. Probe-C/R are the mean accuracy for the correct image-prompt match. Top scores are in \textbf{bold} while second are \underline{underlined}.}
    \resizebox{\columnwidth}{!}{\begin{tabular}{lcccc}
        
        Model & ImageNet & RelComp & Probe-C  & Probe-R  \\
        \hline
        ViLT & -- & 76.00 & 90.78 & 69.00 \\
        BridgeTower & -- & 85.00 & 90.06 & 82.20 \\
        \hline\hline
        FLAVA & 56.83 & 47.12 & 83.85 & 68.29 \\
        CLIP ViT B32 & \textbf{63.60} & 51.93 & 88.15 & 53.52 \\
        \hline
        CLIP Patched (T) & 57.85 & \textbf{67.85 }& 89.49 & \underline{71.14} \\
        CLIP Patched (V) & \underline{61.45} & 54.66 & \underline{89.81} &  61.40\\
        CLIP Patched (VT) & 54.61 & \underline{64.27} & \textbf{90.30} & \textbf{71.20} 
    \end{tabular}}
    \label{tab:training_results2}
\end{table}
Table \ref{tab:patch_ablation} shows the results based on different alphas for RelComp and ImageNet. There is a definite trade-off between original performance and performance on the new task. We also see that training only the text encoder yields the greatest improvement in these tasks but also the largest ``forgetting''. Some examples of where CLIP patched improved over CLIP in Probe-R is shown in Figure \ref{fig:clip_patched}. The first column are the original images, the second the attention maps of visual and text features for CLIP ViT-B/32 and the third are the attention maps for CLIP Patched (VT). The values are the softmax confidence for the correct prompt $P_{R_1}$ shown as $1)$ versus the incorrect prompt $P_{R_2}$ where the predicate is switched $2)$. Similar examples for Probe-C are shown in Figure \ref{fig:clip_patched_comp}. For each group, the first image and its corresponding prompt are on top, and the second image and prompt are on the bottom. The values are the softmax confidence for the corresponding prompt when compared to the alternative prompt.

\section{Model Details}
\label{supp:model_details}
A summary of the model details can be found in Table \ref{tab:models_and_data}. The highest performing model is BridgeTower but it also had the largest number of parameters and the slowest. Additionally, BridgeTower utilizes a pre-trained CLIP visual encoder, improving upon CLIPs performance. All models require image-text pairs, making a greater number of comparisons difficult, especially for downstream tasks like image classification on ImageNet where there are 1000 classes. However, because FLAVA merges dual-stream encoder output prior to cross-encoding, it is easier to extract feature embeddings prior to the cross-encoding for a greater number of comparisons. This however does not utilize its full potential for performance. Figure \ref{fig:clip_teaser} shows examples of how this image-text pair input is a strength for performance in these kinds of tasks. The bottom shows ViLT and how its visual attention changes based on its input while the top shows CLIP which has consistent attention no matter the text, visual input. 
Table \ref{tab:mscoco_flicker} shows the reported results for the selected models and some CLIP models on the MSCOCO \cite{coco} and Flicker \cite{young2014image} datasets. We do see correlation between performance on these datasets and performance on the proposed datasets in this benchmark. This indicates that retrieval tasks on datasets like MSCOCO may be a good indicator of ``understanding'' at a high-level. Code to run these models is 

\section{Dataset Labelling/Preprocessing/Cleaning}
\paragraph{Probe-R}
This dataset is built off of Visual Genome \cite{visualgenome}. This dataset was created by cleaning the annotations/relationship aliases with relations that are specifically an interaction rather than an attribute which was often an erroneous annotation and grouping relations that are the same despite spelling errors. Objects and predicates are additionally cleaned based on spelling errors. Using the extracted (subject, predicate, object) triplets, unlikely relationships are determined if there is no existing combination of an object-predicate pair or subject-object pair. For each image, the ground truth relation is compared to a highly unlikely swap of subject and predicate. There are set of "positive" images that are images with the subject being swapped and no other objects from the original image. There are also a set of "negative" images that are images with the swapped subject and no other objects from the original image. The predicate swapped is based on the predicates that have not been found in the original dataset to be associated with the original subject and therefore are highly unlikely.

\paragraph{Probe-C}  This dataset is built of the COCO Validation 2014 \cite{coco} dataset. Using the NLP library  
NLTK \cite{nltk} and the COCO caption annotations, words are tagged and pairs of adjective and nouns are 
extracted. These pairs are then manually cleaned to ensure the attribute is indeed an adjective and the object is indeed 
an object. Instead of using unlikely combinations, antonyms were manually mapped to each attribute in order to ensure 
that the attribute is not present in the image. For example, if there is a "a small dog", the comparison prompt is 
"a large dog". There are two splits for this dataset. The first is where the composition is swapped with an antonym and the other is where an object is switched. Each dataset has two images and two captions and comparisons are based on how well the 
model can match the captions to the correct images. 

\paragraph{Probe-B} This dataset is built off COCO Validation 2014 \cite{coco} dataset. Segmentation annotations from the original COCO dataset were used for removing background and/or objects. For set 1, for each image, it removes all other objects using either the segmentation(retaining shape cues). These are replaced with either ``black'', ``gray'', ``scene'' or ''noise''. Images are chosen if they do not have overlapping bounding boxes and if their object area is over a threshold  to allow for better visibility, making the task easier. ``Scene'' fillers are extracted from landscape scenery from a subset of the Kaggle Landscape dataset \cite{kaggle_landscape_dataset} and Indoor Scenes Dataset \cite{quattoni2009recognizing}. The subset of 290 scene filler images were selected based on whether there were any objects in the image that are also in the annotations.

\end{document}